\documentclass[sigconf, nonacm]{acmart}

\usepackage{defs_local}
\usepackage{algorithm}
\usepackage[noend]{algpseudocode}
\usepackage{graphicx}
\usepackage{caption}
\usepackage{subcaption}
\usepackage{pgfplots} 
\usepgfplotslibrary{statistics}
\usetikzlibrary{matrix}
\usepgfplotslibrary{groupplots}
\pgfplotsset{compat=newest}
\usepackage{tikzscale}
\usepackage{svg}
\usepackage{multirow}
\usepackage{tabularx,colortbl}
\usepackage{enumitem}

\newcommand{\vk}{\vek{k}}

\newcommand{\vr}{\vek{r}}
\newcommand{\X}[1]{X_{#1}}

\newcommand{\gtX}[1]{\bar{X}_{#1}}
\newcommand{\hatX}[1]{\hat{X}_{#1}}
\newcommand{\RR}{\mathbb{R}} 
\newcommand{\EE}{\mathbb{E}} 
\newcommand{\Pcal}{\mathcal{P}}
\newcommand{\sib}{\text{Sib}}
\newcommand{\cI}{{{\mathcal{I}}}} 
\newcommand\vldbdoi{XX.XX/XXX.XX}
\newcommand\vldbpages{XXX-XXX}
\newcommand\vldbvolume{14}
\newcommand\vldbissue{1}
\newcommand\vldbyear{2020}
\newcommand\vldbauthors{\authors}
\newcommand\vldbtitle{\shorttitle} 
\newcommand\vldbavailabilityurl{http://vldb.org/pvldb/format_vol14.html}

\newcommand\sysname{DeepMVI}
\newcommand\DiscardT{DropCell}

\begin{document}
\title{Missing Value Imputation on Multidimensional Time Series}

\author{Parikshit Bansal}
\affiliation{%
  \institution{IIT Bombay}
}
\email{parikshitb52@gmail.com}

\author{Prathamesh Deshpande}
\affiliation{%
  \institution{IIT Bombay}
}
\email{pratham@cse.iitb.ac.in}

\author{Sunita Sarawagi}
\affiliation{%
  \institution{IIT Bombay}
}
\email{sunita@iitb.ac.in}

\begin{abstract}
We present \sysname, a deep learning method for missing value imputation in multidimensional time-series datasets. Missing values are  commonplace in decision support platforms that aggregate data over long time stretches from disparate sources, whereas reliable data analytics calls for careful handling of missing data. One strategy is imputing the missing values, and a wide variety of algorithms exist spanning simple interpolation, matrix factorization methods like SVD, statistical models like Kalman filters, and recent deep learning methods.  We show that often these provide worse results on aggregate analytics compared to just excluding the missing data.
%

{\color{black}
\sysname\ expresses the distribution of each missing value conditioned on coarse and fine-grained signals 
along a time series, and signals from correlated series at the same time.  Instead of resorting to linearity assumptions of conventional matrix factorization methods, \sysname\ harnesses a flexible deep network  to extract and combine these signals in an end-to-end manner. To prevent over-fitting with high-capacity neural networks, we design a robust parameter training with labeled data created using synthetic missing blocks around available indices. Our neural network uses a modular design with a novel temporal transformer with convolutional features,  
and kernel regression with learned embeddings.} 

Experiments across ten real datasets, five different missing scenarios, comparing seven conventional and three deep learning methods show that \sysname\ is significantly more accurate, reducing error by more than 50\% in more than half the cases, compared to the best existing method.  Although  slower than simpler matrix factorization methods, we justify the increased time overheads by showing that \sysname\ provides significantly more accurate imputation that finally impacts quality of downstream analytics.

\end{abstract}

\maketitle

\begingroup\small\noindent\raggedright\textbf{PVLDB Reference Format:}\\
\vldbauthors. \vldbtitle. PVLDB, \vldbvolume(\vldbissue): \vldbpages, \vldbyear.\\
\href{https://doi.org/\vldbdoi}{doi:\vldbdoi}
\endgroup
\begingroup
\renewcommand\thefootnote{}\footnote{\noindent
This work is licensed under the Creative Commons BY-NC-ND 4.0 International License. Visit \url{https://creativecommons.org/licenses/by-nc-nd/4.0/} to view a copy of this license. For any use beyond those covered by this license, obtain permission by emailing \href{mailto:info@vldb.org}{info@vldb.org}. Copyright is held by the owner/author(s). Publication rights licensed to the VLDB Endowment. \\
\raggedright Proceedings of the VLDB Endowment, Vol. \vldbvolume, No. \vldbissue\ %
ISSN 2150-8097. \\
\href{https://doi.org/\vldbdoi}{doi:\vldbdoi} \\
}\addtocounter{footnote}{-1}\endgroup

\ifdefempty{\vldbavailabilityurl}{}{
\vspace{.3cm}
\begingroup\small\noindent\raggedright\textbf{PVLDB Availability Tag:}\\
The source code of this research paper has been made publicly available at \url{\vldbavailabilityurl}.
\endgroup
}

\section{Introduction}
In this paper we present a system for imputing missing values across multiple time series occurring in multidimensional databases.  
Examples of such data include sensor recordings along time of different types of  IoT devices at different locations,  daily traffic logs of web pages from various device types and regions, and demand along time for products at different stores.
Missing values are commonplace in analytical systems that integrate data from multiple sources over long periods of time.  Data may be missing because of errors or breakdowns at various stages of the data collection pipeline ranging from faulty recording devices to deliberate obfuscation.  Analysis on such incomplete data may yield biased results misinforming data interpretation and downstream decision making. Therefore, missing value imputation is an essential tool in any analytical systems~\cite{Cambronero2017,milo2020automating,kandel2012profiler,Mayfield2010}. 

Many techniques exist for imputing missing values in time-series datasets including several matrix factorization techniques~\cite{yu2016temporal,troyanskaya2001missing,khayati2019scalable,mei2017nonnegative,mazumder2010spectral,cai2010singular},  statistical temporal models~\cite{li2009dynammo}, and recent deep learning methods~\cite{cao2018brits,fortuin2020gp}. Unfortunately, even the best of existing techniques still incur high imputation errors.  We show that top-level aggregates used in analytics could get worse after imputation with existing methods, compared to discarding missing data parts before aggregation.  Inspired by the recent success of deep learning in other data analytical tasks like entity matching, entity extraction, and time series forecasting, we investigate if better deep learning architectures can reduce this gap for the missing value imputation task.

 The pattern of missing blocks in a time series dataset can be quite arbitrary and varied.  Also, datasets could exhibit very different characteristics in terms of the length and number of series,  amount of repetitions (seasonality) in a series, and correlations across series. An entire contiguous block of entries might be missing within a time series, and/or across multiple time-series. The signals from the rest of the dataset that are most useful for imputing a missing block would depend on the size of the block, its position relative to other missing blocks, patterns within a series, and correlation (if any) with other series in the dataset.  If a single entry is missing, interpolation with immediate neighbors might be useful. If a range of values within a single time series is missing, repeated patterns within the series and trends from correlated series might be useful. If the same time range across several series is missing, only patterns within a series will be useful.
 
 Existing methods based on matrix factorization can exploit across series correlations but are not as effective in combining them with temporal patterns within a series.  
 Modern deep learning methods, because of their higher capacity and flexibility, can in principle combine diverse signals when trained end to end.  However,  designing a neural architecture whose parameters can be trained accurately and scalably across diverse datasets and missing patterns proved to be non-trivial.  Early solutions based on popular architectures for sequence data, such as recurrent neural networks (RNNs) have been shown to be worse both in terms of accuracy and running time.   We explored a number of alternative architectures spanning Transformers, CNNs, and Kernel methods. A challenge we faced when training a network to combine a disparate set of potentially useful signals was that, the network was quick to overfit on easy signals. Whereas robust imputation requires that the network harness all available signals.  
 After several iterations, we converged on a model and training procedure, that we call \sysname\ that is particularly suited to the missing value imputation task.



\begin{figure}[h]
    \centering
    \includegraphics[width=0.5\textwidth]{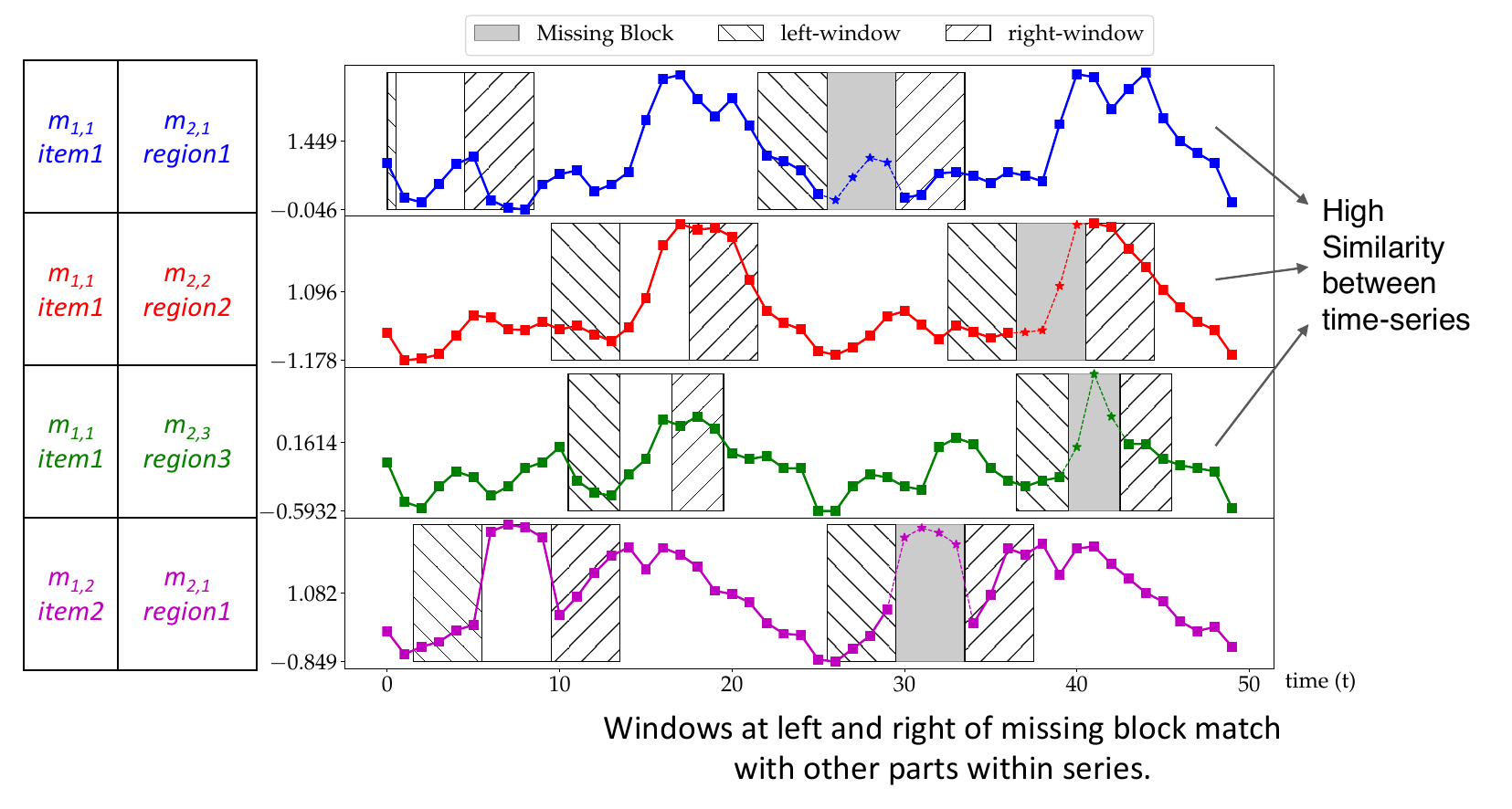}
    \caption{Grey-shaded regions denote missing blocks. Patterns of left and right windows around each missing block match with another part of the same series. Series 1-3 have high similarity and series 1,2,4 show good window match along time.  
    }
    \label{fig:missing_patterns}
\end{figure}

\subsection{Contributions}
{\color{black}
(1) We propose a tractable model to express each missing value using a distribution conditioned on available values at other times within the series and values of similar series at the same time.  (2) We design a flexible and modular neural architecture to extract fine-grained, coarse-grained, and cross-series signals to parameterize this distribution. (3) We provide a robust method of training generalizable parameters  by simulating missing patterns around available indices that are identically distributed to the actual missing entries. (4) Our neural architecture includes a Temporal transformer that differs from off-the-shelf  Transformers in our method of creating contextual keys used for self-attention. 
(5) 
We propose the use of Kernel regression for incorporating information from correlated time-series. 
%
This method of extracting relatedness is scalable and extends naturally to multidimensional datasets, that none of the existing methods handle.
(6) We achieve 20--70\% reduction in imputation error as shown via an extensive comparison with both state of the art neural approaches and traditional approaches across ten real-life datasets and five different patterns of missing values.  We show that this also translates to more accurate aggregates analytics on datasets with missing entries.
(7) Our method is six times faster than using off-the-shelf deep-learning components for MVI.}

\section{Preliminaries and Related Work}
We present a formal problem statement, discuss related work, and provide background on relevant neural sequence models.
\subsection{Problem Statement}
{\color{black}
We denote our multidimensional time series dataset as an  $n+1$ dimensional data tensor of real values $X \in \RR^{n+1}$.  The dimensions of $X$ are denoted as ($K_1$,$K_2$,...,$K_n$, $K_{n+1}$).  The dimension $K_{n+1}$ denotes a regularly spaced time index which, without loss of generality we denote as $\{1,\ldots,T\}$. 
Each  $K_i$ is a dimension  comprising of a discrete set of members $\{m_{i,1},\ldots,m_{i,|K_i|}\}$.  Each member $m_{ij}$ could be either a categorical string or a real-valued vector.
For example, a retail sales data might consist of two such dimensions: $K_1$ comprising of categorical members denoting identity of items sold and $K_2$ comprising of stores 
where a store is defined in terms of its continuous latitude and longitude value.}  
We denote a specific combination of members of each dimension as $\vk=k_1,\ldots,k_n$ where each $k_i \in \mathrm{Dim}(K_i)$.  We refer to the value at a combination $\vk$ and time $t$ as $\X{\vk,t}$.  For example in Figure~\ref{fig:missing_patterns} we show four series of length 50 and their index $\vk$ sampled from a two dimensional categorical space of item-ids and region-ids.
%
We are given an $X$ with some fraction of values  $\X{\vk,t}$ missing.  Let $M$ and $A$ be tensors of only ones and zeros with same shape as $X$ that denote the missing and available values respectively in $X$.
We use $\cI(M)$ to denote all missing values' $(\vk,t)$ indices. 
The  patterns in $\cI(M)$ of missing index combinations can be quite varied --- for example missing values may be in contiguous blocks or isolated points; across time-series the missing time-ranges may be overlapping or missing at random; or in an extreme case called Blackout a time range may be missing in all series.  
Our goal is to design a procedure that can impute the missing values at the given indices $I(M)$  so that the error between the imputed values $\hatX{}$ and ground truth values $\gtX{}$ is minimized.
\begin{equation}
    \sum_{(\vk,t)\in \cI(M)} \cE(\hatX{\vk,t}, \gtX{\vk,t})
\end{equation}
where $\cE$ denotes error functions such as root mean square error (RMSE) and mean absolute error (MAE).  As motivated in Figure~\ref{fig:missing_patterns} both patterns within and across a time series may be required to fill a missing block.

\subsection{Related Work}
Missing value imputation in time series is an age-old problem~\cite{little2002single}, with several solutions that we categorize into matrix-completion methods,  conventional statistical time-series models, and recent deep learning methods (discussed in Section~\ref{sec:relate:deep}).  However, all these prior methods are for single-dimensional series.  So, we will assume $n=1$ for the discussions below. 
\myparagraph{Matrix completion methods}
These methods~\cite{yu2016temporal,troyanskaya2001missing,khayati2019scalable,mei2017nonnegative,mazumder2010spectral,cai2010singular}, view the time-series dataset as a matrix $X$ with rows corresponding to series and columns corresponding to time.  They then apply various dimensional reduction techniques to decompose the matrix as $X\approx UV^T$ where $U$ and $V$ represent low-dimensional embeddings of series  and time respectively. The missing entry in a series $i$ and position $t$ is obtained by multiplying the corresponding embeddings. 
A common tool is the classical Singular Value Decomposition (SVD) and this forms the basis of three earlier techniques: SVDImp\cite{troyanskaya2001missing}, SoftImpute~\cite{mazumder2010spectral}, and SVT~\cite{cai2010singular}.  All these methods are surpassed by a recently proposed centroid decomposition (CD) algorithm called 
CDRec\cite{khayati2019scalable}.  
CDRec performs recovery by first using interpolation/extrapolation to initialize the missing values. Second, it computes the CD and keeps only the first k columns of U and V, producing $U_k$ and $V_k$, respectively. Lastly, it imputes values using $X = U_k V_k^T$. This process iterates until the normalized Frobenius norm between the matrices before and after the update reaches a small threshold. 

A limitation of pure matrix decomposition based methods is that they do not capture any dependencies along time. 
TRMF\cite{yu2016temporal} proposes to address this limitation by introducing a regularization on the temporal embeddings $V$ so that these confirm to auto-regressive structures commonly observed in time-series data.  STMVL is another algorithm that smooths along time and is designed to recover missing values in spatio-temporal data using collaborative filtering methods for matrix completion.  





\myparagraph{Statistical time-series models}
DynaMMO\cite{li2009dynammo},
is an algorithm that creates groups of a few time series based on similarities that capture co-evolving patterns. They fit a Kalman Filter model on the group using the Expectation Maximization (EM) algorithm. The Kalman Filter uses the data that contains missing blocks together with a reference time series to estimate the current state of the missing blocks. The recovery is performed as a multi-step process. At each step, the EM method predicts the value of the current state and then two estimators refine the predicted values of the given state, maximizing a likelihood function.

{\color{black}
\myparagraph{Pattern Based Methods}
TKCM \cite{wellenzohn2017continuous} identifies and uses repeating patterns (seasonality) in the time series’ history. They find similarity between window of measures spanning all time series and window around the query time index using Pearson's correlation coefficient and do 1-1 imputation using the mean value of the matched blocks. Though promising this method performs poorly compared to other baselines like CDRec, on each dataset \cite{khayati2020mind}, hence we have excluded it from our analysis. Deep Learning architectures have been shown to perform better at query-pattern search and corresponding weighted imputation \cite{Vaswani2017} which we exploit in our work.
}
We present an empirical comparison with SVDImp (as a representative of pure SVD methods), CDRec, TRMF, STMVL, and DynaMMO and show that our method is significantly more accurate than all of them.


\subsection{Background on Neural Sequence Models}
We review\footnote{Readers familiar with Deep Learning may skip this subsection.} two popular neural architectures for processing sequence data. 
\subsubsection{Bidirectional Recurrent Neural Networks}

\newcommand{\vU}{\vek{U}}
\newcommand{\vb}{\vek{b}}
Bidirectional RNN \cite{graves2005framewise} is a special type of RNN that captures dependencies in a sequence in both forward and backward directions. Unlike forecasting, context in both forward and backward directions is available in MVI task. 
%
%
Bidirectional RNN maintains two sets of parameters, one for forward and another for backward direction. Given a sequence $X$, the forward RNN maintains a state $\vh^f_t$ summarizing $X_1\ldots X_{t-1}$, and backward RNN maintains a state $\vh^b_t$ summarizing $X_T\ldots X_{t+1}$. These two states jointly can be used to predict a missing value at $t$.  
%
Because each RNN models the dependency along only one direction, a bidirectional RNN can compute loss at each term in the input during training. 



\subsubsection{Transformers} 
\label{sec:Transfomers}
A Transformer \cite{Vaswani2017} is a special type of feed-forward neural network that captures sequential dependencies through a combination of self-attention and feed-forward layers. Transformers are primarily used on text data for language modelling and various other NLP tasks~\cite{devlin2018bert}, but have recently also been used for time-series forecasting \cite{li2019enhancing}.

Given an input sequence $X$ of length $T$, a transformer processes it as follows:  It first embeds the input $X_{t}$ for each  $t \in [1, T]$ into a vector $E_t \in \RR^{p}$, called the input embedding. It also creates a positional encoding vector at position $t$ denoted as $e_t \in \RR^p$. 
\begin{align}
    e_{t,r}=
    \begin{cases}
        \sin(t/10000^{\frac{r}{p}}), & \textrm{if}~~~ r\%2 == 0 \\
        \cos(t/10000^{\frac{r-1}{p}}), & \textrm{if}~~~ (r-1)\%2 == 0
    \end{cases}
    \label{eqn:position_encoding}
\end{align}
Then it uses linear transformation of input embedding and positional encoding vector to create query, key, and value vectors. 
\begin{align}
\label{eq:oldKey}
    Q_t &= (E_t + e_t) W^Q \quad K_t = (E_t + e_t) W^K \quad V_t = (E_t + e_t) W^V
\end{align}
where the $W$s denote trained parameters.
Key and Value vectors at all times $t \in [1, T]$ are stacked to create matrices $K$ and $V$ respectively. Then the query vector at time $t$ and keys pair at other positions $t' \neq t$ are used to compute a self-attention distribution, which is used to compute a vector at each $t$ as an attention weighted sum of its neighbors as follows:
\begin{align}
    \vh_t = \textrm{Softmax}(\frac{Q_tK^T}{\sqrt{p}})V
    \label{eqn:vanilla_transformer_attn}
\end{align}
Such self-attention can capture the dependencies between various positions of the input sequence. Transformers use multiple such self-attentions to capture different kinds of dependencies, and these are jointly referred as multi-headed attention.  
In general, multiple such layers of self-attention are stacked.  The final vector $\vh_t$ at each $t$ presents a contextual representation of each $t$.

For training the parameters of the transformer, a portion of the input would be masked (replaced by 0). We denote the masked indices by $M$. 
The training loss is computed only on the masked indices in $M$. This is because multiple layers of self-attention can compute $\vh_t$ as a function of any of the input values. This is unlike bidirectional RNNs where the forward and backward RNN states clearly demarcate the values used in each state.  This allow loss to be computed at each input position.  However, transformers are otherwise faster to train in parallel unlike RNNs.  

\subsection{Related work in Deep-learning}
\label{sec:relate:deep}
In spite of the recent popularity and success of deep learning (DL) in several difficult tasks, existing work on the MVI task are few in number. Also there is limited evidence of DL methods surpassing conventional methods across the board.
MRNN\cite{yoon2018estimating} is one of the earliest deep learning methods. MRNNs use 
Bidirectional RNNs to capture  context of a missing block within a series, and capture correlations across series using a fully connected network.  However, a detailed empirical evaluation in \cite{khayati2020mind} found MRNN to be orders of magnitude slower than above matrix completion methods, and also (surprisingly) much worse in accuracy.
More recently, BRITS\cite{cao2018brits}  is another method that also uses bidirectional RNNs. At each time step $t$ the RNN is input a column $X_{:,t}$ of $X$.  The RNN state is the black box charged with capturing both the dependencies across time and across series. %
%
GP-VAE\cite{fortuin2020gp} adds more structure to the dependency by first converting each data column $X_{:,t}$ of $X$ to a low-dimensional embedding, and then using a Gaussian Process to capture dependency along time in the embedding space.  Training is via an elaborate structured variational method.  On record time series datasets GP-VAE has been shown to be worse empirically than BRITS, and seems to be geared towards image datasets. 
\nocite{liu2019naomi} 

%
Compared to these deep models, our network architecture is more modular and light-weight in design, allows for more stable training without dataset specific hyper-parameter tuning, more accurate, and significantly faster.

\myparagraph{Other Deep Temporal Models}
Much of the work on modeling time series data has been in the context of the forecasting task.  State of the art methods for forecasting are still RNN-based~\cite{FlunkertSG17,deshpande2019streaming,salinas2019high,sen2019think}.  The only exceptions is \cite{li2019enhancing} that uses 
convolution to extract local context features of the time-series and then a transformer to capture longer-range features.  Such
transformer and convolutions models have been quite successful in speech transcription literature \cite{li2019jasper}. 
Our architecture is also based on transformers and convolutions but our design of the keys and queries is better suited for missing value imputation.  Further, we also include a fine-grained context and a second kernel regression model to handle across time correlations. 

\begin{figure}
    \centering
    \includegraphics[width=0.5\textwidth]{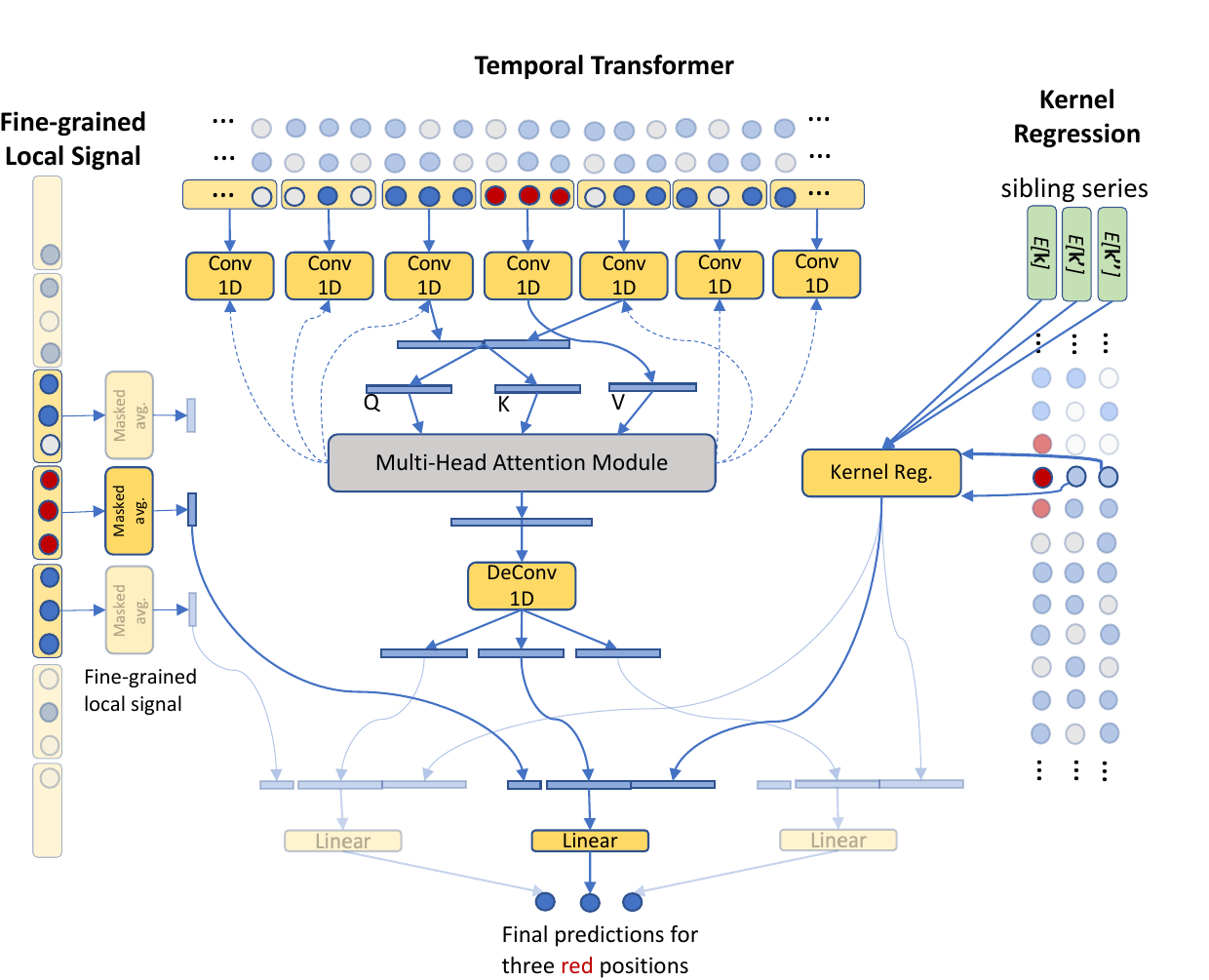}
    \caption{Architecture of \sysname. Here model is shown imputing the three circles marked in {\color{red} red} at the top. The temporal transformer convolves on window of size $w=3$ to create queries, keys and values for the multi-headed attention. The deconvolution creates three vectors, one for each red circle. These are concatenated with fine-grained signal and kernel regression to predict the final output.}
    \label{alg:trsf-arch}
\end{figure}

\section{\sysname: The conceptual Model }
\label{sec:ctrain}
{\color{black}
We cast the missing value imputation task as solving an objective of the form:

$$
\mathrm{max}_{\hatX{}} \prod_{(\vk,t)\in \cI(M)}  \Pr(\hatX{\vk,t}|X,A;\theta) 
$$
where $\theta$ are the parameters of the model and $A$ is the mask denoting available values in $X$. 
%
$X$ is the entire set of available values, and any tractable models will need to break-down the influence of $X$ at an $(\vk,t)$ into simpler, learnable subparts.  
State of the art deep learning methods
such as BRITS, simplify the dependence as:
$$
\Pr(\hatX{\vk,t}|\X{},A;\theta ) = \Pr(\hatX{\vk,t}|\X{\bullet,1\ldots t-1}, \X{\bullet,t+1\ldots T}, \theta)
$$
The first part $\X{\bullet,1\ldots t-1}$ denotes the entire vector of values over all times before $t$ and likewise $\X{\bullet,t+1\ldots T}$ for after $t$.  Note here 
observed values at time $t$ from correlated sequences are ignored. Also, each of these sequences are summarized using RNNs that take as input values over all series $\X{\bullet,j}$ at each step $j$. This limits the scalability on the number of series.

\newcommand{\XA}[1]{[X,A]_{#1}}
In contrast, DeepMVI captures the dependence both along time within series $\vk$ and along other related series at time $t$ to simplify the dependency structure as:  
\begin{align}
\label{eq:concept} \Pr(\hatX{\vk,t}|X,A;\theta) = \Pr(\hatX{\vk,t} |
 \X{\vk,\bullet},
 \X{\sib(\vk),t}, A, \theta)
\end{align}
The first part $\X{\vk,\bullet}$ is used to capture the long term dependency within the series $\vk$ and also fine-grained signals from the immediate temporal neighborhood of $t$.  The second part extracts signals 
from related series $\sib(\vk)$ at time $t$. The notion of $\sib(\vk)$ is defined using learnable kernels that we discuss in Section~\ref{ssec:kr}.  Unlike conventional matrix factorization or statistical methods like Kalman Filters that assume fixed  functional forms, we depend on the universal approximation power of neural networks to extract signals from the context $\X{\vk,\bullet}$ and  $\X{\sib(\vk),t}$ to create a distribution over the  missing value at $\vk,t$. The neural network architecture we used for parameterizing this distribution is described Section~\ref{sec:mviDL}.

The parameters ($\theta$) of high-capacity neural networks need to be trained carefully to avoid over-fitting.  We do not have separate labeled datasets for training the model parameters.   Instead, we need to create our own labeled dataset using available values $A$ in the same data matrix $X$.
We  create a labeled dataset from randomly sampled $(\vk_i,t_i)$ indices  from the available set $A$.  Each index $(\vk_i,t_i)$ defines a training instance with input  $\vx_i=(\X{\vk,\bullet},\X{Sib(\vk),t},A)$ and continuous output $y_i=\X{\vk_i,t_i}$ as label, and thus can be cast as a standard regression model.   In order for the trained parameters $\theta$ to generalize to the missing indices in $\cI(M)$, the available values in the context of a $(\vk_i,t_i)$ used in training need to be distributed identically to those in $\cI(M)$. We achieve this by creating synthetic missing values around each $(\vk_i,t_i)$. The shape of the missing block is chosen by sampling a shape $B_i$ from anywhere in $M$.  Note the shape $B_i$ is a cuboid characterized by just the {\em number} (and not the position) of missing values along each of the $n+1$ dimensions.  We then place $B_i$ randomly around   $(\vk_i,t_i)$, to create a new availability matrix $A_i$ after masking out the newly created missing entries around $(\vk_i,t_i)$.  Our training objective thereafter is simple likelihood maximization over the training instances as follows:
\begin{equation*}
\theta^*=    \mathrm{argmax}_{\theta} \underset{(\vk_i,t_i) \in \cI(A)}{\sum}[\log\Pr(\X{\vk,t}|\X{\vk,\bullet},\X{\sib(\vk),t}, A^i,\theta)]
\end{equation*}
Thus $\theta^*$ has been trained to predict the true value of $|A|$ instances, where our method of sampling $A^i$ ensures that these are identically distributed as the missing entries $\cI(M)$. We further prevent over-fitting on the training instances by using a validation dataset to do early stopping.  This implies that we can invoke the standard ML guarantees of generalization to claim that our model will generalize well to the unseen indices.
}
\section{\sysname: The Neural Architecture}
We implement the conditional model of Equation~\ref{eq:concept} as a multi-layered modular neural network. The first module is the temporal transformer that takes as input $\X{\vk,\bullet}$ and extracts two types of imputation signals along time:
longer-term seasonality represented as a output vector $\vh^{\mathrm{tt}} = TT_\theta(\X{\vk,\bullet}, A_{\vk,\bullet})$, and a fine-grained signal from the immediate neighborhood of $(\vk,t)$ represented as $\vh^{\mathrm{fg}} = FG_{\theta}(\X{\vk,\bullet}, A_{\vk,\bullet})$. 
The second module is the kernel regression that extracts information from related series at time $t$ i.e., from $\X{\sib(\vk),t}$ to output another hidden vector $\vh^{\mathrm{kr}} = KR_{\theta}(\X{\bullet,t}, A{\bullet,t})$. The last layer combines these three outputs as a light-weight linear layer to output a mean value of the distribution of $\hatX{\vk,t}$ as follows:\begin{align}
    \mu[\hatX{\vk,t}] &= \bm{w}_o^T [\vh^{\mathrm{tt}},\vh^{\mathrm{fg}}, \vh^{\mathrm{kr}}] + \bm{b}_o
\end{align}
The above mean is used to model a Gaussian distribution for the conditional probability with a shared variance.
A pictorial depiction of our pipeline appears in Figure~\ref{alg:trsf-arch}. We describe each of these modules in the following sections.

\subsection{Temporal Transformer Module}
\label{subsec:tt}
We build this module for capturing temporal dependency in the series akin to seasonality and draw inspiration from the Transformer~\cite{Vaswani2017} architecture.
The parallel processing framework of attention module provides a natural formulation for handling missing values by masking inputs in contrast to the sequential modeling in RNNs.
However, our initial attempts at using the vanilla Transformer model (described in Sec~\ref{sec:Transfomers}) for the MVI task was subject to over-fitting on long time-series, and inaccurate for block missing values. We designed a new transformer specifically suited for the MVI task which we call the Temporal Transformer. 

With the slight abuse of notation we override the definition of data and availability tensors $X$ and $A$ to 1-dimensional data and availability series respectively. Accordingly $I(A)$ and $I(M)$ are also overridden to confirm to series data. 
We use $I(A) = I-I(M)$ to denote all the indices that are not missing in $X$. 
We next describe how Temporal Transformer computes the function $TT_{\theta}(X,A)$.

\myparagraph{Window-based Feature Extraction}
A linear operation on the window $\X{jw:(j+1)w}$ computes a $p$-dimensional vector $Y_j$ as follows:
\begin{align}
\label{eqn:TT_Yj} Y_j = W_f \X{jw:(j+1)w} + b_f
\end{align}
where $W_f \in \RR^{p\times w}$ and $b_f \in \RR^{p}$ are parameters. This operation is also termed as non-overlapping convolutions in Deep Learning literature. 

We use self-attention on $Y_j$ vectors obtained from Eqn.~\ref{eqn:TT_Yj} above. We now describe the computation of query, key and value vectors for the self-attention.
\myparagraph{Query, Key and Value functions}
For an index $j$ (corresponding to the vector $Y_j$), we define the functions query and key functions $Q(\cdot)$ and $K(\cdot)$ as the functions of $Y_{j-1}$ and $Y_{j+1}$ respectively. Similarly we define value $V(\cdot)$ as function of $Y_j$.  
\begin{align}
\label{eqn:TT_query} Q(Y,j) &= ([Y_{j-1},Y_{j+1}]+e_j)W_q + b_q \\ 
\label{eqn:TT_key} K(Y,j, A) &= (([Y_{j-1},Y_{j+1}]+e_j)W_k + b_k)\cdot \prod_{i=jw}^{(j+1)w}A_{i}\\
\label{eqn:TT_value} V(Y_j) &= Y_{j}W_v + b_v
\end{align}
where $W_q,W_k \in \RR^{2p \times 2p},W_v \in \RR^{p \times p}$, $e_j$ is the positional encoding of index $j$ defined in Eqn.~\ref{eqn:position_encoding}. The product of $A_i$-s in Eqn.~\ref{eqn:TT_key} is one only if all values in the window are available. This prevents attention on windows with missing values. Note that keys and values are calculated for other indices $j' \neq j$ as well. 

\myparagraph{Attention Module}
Attention module computes the attention-weighted sum of vectors $V(Y_{\bullet})$. The attention-weighted sum at index $j$ is calculated as follows:
\begin{align}
    \label{eqn:TT_attn}
    \mathrm{Attn}(Q(\cdot), K(\cdot), V(\cdot), A, j) &= \frac{\sum_{j'}\langle Q(Y,j), K(Y,j',A) \rangle V(Y_{j'})}{\sum_{j'} \langle Q(Y,j), K(Y,j',A) \rangle }
\end{align}
Note that for indices $j'$ with missing values, including the index $j$, the key $K(Y, j', A)$ is zero. Hence such indices are not considered in the attention. 

\myparagraph{MultiHead Attention}
Instead of using only one attention module, our temporal transformer uses multiple instantiations of the functions $Q(Y, j)$, $K(Y, j, A)$, $V(Y_j)$. We compute $n_{\mathrm{head}}$ such instantiations and denote them using index $l=1\ldots n_{\mathrm{head}}$.

We obtain the output of multi-head attention by concatenating the output vectors of all $n_{\mathrm{head}}$ attentions (obtained from Eqn.~\ref{eqn:TT_attn}) into a vector $\vh \in \RR^{pn_{\mathrm{head}}}$.

\begin{align}
    \vh_j = [\mathrm{Attn}^1(\cdots), \ldots, \mathrm{Attn}^{n_{\mathrm{head}}}(\cdots)]
\end{align}

\myparagraph{Decoding Attention Output}

Vector $\vh_j$ is the output vector for window $X_{jw:(j+1)w}$. Decoding module first passes the vector $\vh_j$ through a feed-forward network to obtain the vector $\vh_j^{\mathrm{ff}}$. It then transforms this vector to obtain the output vectors for positions $\{jw,\ldots,t,\ldots,(j+1)w \}$:

\begin{align}
    \label{eqn:TT_decodeMLP} \vh_j^{\mathrm{ff}} &= \mathrm{ReLU}(W_{d_2}(\mathrm{ReLU}(W_{d_1}(\mathrm{ReLU}(\vh_j)))))\\
    \label{eqn:TT_deconv} \vh_j^{\mathrm{tt}} &= \mathrm{ReLU}(W_d\vh_j^{\mathrm{ff}} + b_d)
\end{align}
where $W_d \in \RR^{w\times p \times p}$. Note that $\vh_j^{\mathrm{tt}} \in \RR^{w \times \bullet}$ consists of output vectors for all indices in the $j$-th window. We can obtain the output vector corresponding to index $t$ as $\vh^{\mathrm{tt}} = \vh_j^{\mathrm{tt}}[t\%w]$ as the final output of the $TT_{\theta}(X,A,t)$ module.


\subsubsection{Fine Grained Attention Module}
\label{ssec:fg}
We utilise this module to capture the local structure from immediate neighbouring time indices which are especially significant in the case of point missing values. Let the time index $t$ be part of the window $j=\big\lfloor \frac{t}{w} \big\rfloor$ with start and end times as $t_s^j$ and $t_e^j$ respectively. Then we define the function $FG_{\theta}(X, A)$ as

\begin{align}
\label{eqn:finegrained}
FG_{\theta}(X, A) = \vh^{\mathrm{fg}} =  \frac{\underset{j \in I(A)}{\sum} \X{j}}{|I(A)|}
\end{align}

\subsection{Kernel Regression Module}
\label{ssec:kr}
We build the kernel regression module to exploit information from correlated series along each of the $n$ data dimensions.  A series in our case is associated with  $n$ variables $\vk=k_1,\ldots, k_n$.  

{\color{black}
\myparagraph{Index Embeddings}
First, we embed each dimension member in a space that preserves their relatedness.  If a member $m_{ij}$ of a dimension $K_i$ is categorical we learn an embedding vector $E_\theta(m_{ij}) \in R^{d_i}$.   When the dimension is real-valued, i.e., $m_{ij} \in R^p$, we use $E_\theta(m_{ij})$ to denote any feed-forward neural network to transform the raw vector into a $d_i$-dimensional vector.}

We define relatedness among series pairs.  We only consider series pairs that differ in exactly one dimension.  We call these sibling series:
\myparagraph{Defining Siblings} 
We define siblings for an index $\vk$ along dimension $i$, $\sib(\vk,i)$ as the set of all 
indices $\vk'$ such that $\vk$ and $\vk'$ differ only at $i$-th dimension.
\begin{align}
    \sib(\vk,i) = \{ \vk' :k'_j = k_j ~~ \forall j \neq i \land k'_i \neq k_i \}
    \label{eqn:siblings}
\end{align}
Here we override the notation $\sib(\vk)$ (used earlier in Eqn.~\ref{eq:concept}) to identify the siblings along each dimension. For example, in a retail sales data, that contains three items \{$i_0, i_1, i_2$\} and four regions \{$r_0, r_1, r_2, r_3$\}, siblings of an (item, region) pair $\vk=(i_1, r_2)$ along the product dimension would be 
$\sib(\vk,0) = \{(i_0,r_2), (i_2,r_2)\}$
and along the region dimension would be $\sib(\vk,1) = \{(i_1,r_0), (i_1,r_1), (i_1,r_3)\}$.
\myparagraph{Regression along each dimension}
An RBF Kernel computes the similarity score $\mathcal{K}(k_i,k'_i)$ between indices $k_i$ and $k'_i$ in the $i$-th dimension:

\begin{align}
    \mathcal{K}(k_i,k'_i) = \exp \Big(-\gamma*||E[k_i] - E[k'_i]||_2^2 \Big)
\end{align}

Given a series $X$ at index $(\vk,t)$, for each dimension $i$, we compute the kernel-weighted sum of measure values as
\begin{align}
\label{eqn:KR_U} U_{(\vk,i),t} = \frac{\sum_{\vk' \in  \sib(\vk,i)}X_{\vk',t} \mathcal{K}(k_i,k'_i) A_{\vk',t}}{\sum_{\vk' \in  \sib(\vk,i)}\mathcal{K}(k_i,k'_i) A_{\vk',t}}
\end{align}
where $A_{\vk',t} = 1$ for non-missing indices and $0$ for missing indices. 

When a dimension $i$ is large, we make the above computation efficient by pre-selecting the top $L$ members based on their kernel similarity.
%

We also compute two other measures: Sum of kernel weights and the variance in $X$ values along each sibling dimension:
\begin{align}
    \label{eqn:KR_W} W_{(\vk,i),t} &= \sum_{\vk' \in  \sib(\vk,i)}\mathcal{K}(k_i,k'_i)A_{\vk',t} \\
    \label{eqn:KR_V} V_{(\vk,i),t} &= Var(X_{\sib(\vk, i),t})
\end{align}

The last layer of the  kernel-regression module is concatenation of $U$, $V$, and $W$ components:
\begin{align}
    \vh^{\mathrm{kr}} = \mathrm{Concat}(U_{(\vk,i),t}, V_{(\vk,i),t}, W_{(\vk,i),t})
\end{align}
where $\vh^{\mathrm{kr}} \in \RR^{3n}$.

\begin{figure}
\begin{algorithmic}[1]
\Procedure \sysname{$X, A, M$}
\State TrainData=$(\vk_i,t_i) \in A, A^i$=random misses around $(\vk_i,t_i)$.
\State model $\gets \mathrm{CreateModel}()$ \texttt{/* Sec.~\ref{para:net_default_params} */}

\For {$\mathrm{iter}=0$ to $\mathrm{MaxIter}$}
\For{\textbf{each} $(\vk_i,t_i,A^i)\sim \mathrm{Batch}$(TrainData)}
\State {$\hat{X}_{\vk_i, t_i} \gets \mathrm{ForwardPass}(X, A^{i}, \vk_i, t_i)$}
\EndFor
\State Update model parameters $\Theta$.
\State Evaluate validation data for early stopping.
\EndFor

\State \texttt{/* Impute test-blocks */}
\State{$\hat{X} \gets \mathrm{ForwardPass}(X, A, \bullet, \bullet)$ over all test blocks in $\cI(M)$}. \\
\Return $\hat{X}$
\EndProcedure

\Procedure{ForwardPass}{$X$, $A$, $\vk$, $t$}
\State $\vh^{\mathrm{tt}}, \vh^{\mathrm{fg}} = TT(X, A)$.
\State $\vh^{\mathrm{kr}} = KR(X, A)$. \texttt{Section \ref{ssec:kr}} \\
\Return $\vh^{\mathrm{tt}}$, $\vh^{\mathrm{fg}}$, $\vh^{\mathrm{kr}}$
\EndProcedure

\Procedure{TT}{$X$, $A$}
\State Index of the block containing time $t$ is $j = T \% w$.
\State $Y_j = W_f X_{jw:(j+1)w} + b_f$.
\State \texttt{/*Similarly compute $Y_{j-1}$ and $Y_{j+1}.$*/}
\State Compute Query, Keys, and Values using Equations \ref{eqn:TT_query}, \ref{eqn:TT_key}, \ref{eqn:TT_value}.
\State Calculate $\mathrm{Attn(Q(\cdot), K(\cdot), V(\cdot), A, j)}$ using Eqn.~\ref{eqn:TT_attn}.
\State Calculate multi-head attention:
\begin{align*}
    \vh_j = [\mathrm{Attn}^1(\cdots), \ldots, \mathrm{Attn}^{n_{\mathrm{head}}}(\cdots)]
\end{align*}
\State Compute vector $\vh_j^{\mathrm{tt}}$ using Equations \ref{eqn:TT_decodeMLP} and \ref{eqn:TT_deconv}.
\State $\vh^{\mathrm{tt}} = \vh_j^{\mathrm{tt}}[t\%w]$.
\State Compute the fine grained attention vector:
\begin{align*}
    \vh^{\mathrm{fg}} = FG_{\theta}(X, A) \texttt{/* Eqn~\ref{eqn:finegrained} */}
\end{align*}
\Return $\vh^{\mathrm{tt}}$, $\vh^{\mathrm{fg}}$
\EndProcedure

\Procedure{KR}{$X$, $A$}
\State Compute the vectors $U_{\bullet}$, $W_{\bullet}$, and $V_{\bullet}$ (Equations \ref{eqn:KR_U}, \ref{eqn:KR_W}, \ref{eqn:KR_V}).
\State $\vh^{\mathrm{kr}} = \mathrm{Concat}(U_{(\vk,i),t}, V_{(\vk,i),t}, W_{(\vk,i),t})$. \\
\Return $\vh^{\mathrm{kr}}$
\EndProcedure
\end{algorithmic}
\caption{The \sysname\ training and imputation algorithm}
\label{alg:pcode_new}
\end{figure}

\newcounter{paranumbers}
\newcommand\paranumber{\stepcounter{paranumbers}\arabic{paranumbers}}


\subsection{Network Parameters and Hyper-parameters}
\label{sec:train}
The parameters of the network span the temporal transformer, the embeddings of members of dimensions used in the kernel regression, and the parameters of the output layer. We use $\Theta$ to denote all the trainable parameters in all modules: $$\Theta = \{ W_f, b_f, W_q, b_q, W_k, b_k, W_v, b_v, W_{d_1}, W_{d_2}, W_d, b_d,  \bm{w}_o, \bm{b}_o, E[m_{\bullet,\bullet}] \}.$$ 
These parameters are trained using the training objective described in Section~\ref{sec:ctrain} on the available data. Any off-the-shelf stochastic gradient method can be used for solving this objective. We used Adam with a learning rate 1e-3. 

{\color{black}
\label{para:net_default_params}
\paragraph{Network Hyper-parameters:} 
Like any deep learning method, our network also has hyper-parameters that control the size of the network, which in turn impacts  accuracy in non-monotonic ways.  Many techniques exist for automatically searching for optimal values of hyper-parameters~\cite{citeHyper} based on performance on a validation set.  These techniques are applicable in our model too but we refrained from using those for two reasons: (1) they tend to be computationally expensive, and (2) we obtained impressive results compared to almost all existing methods in more than 50 settings without 
extensive dataset specific hyper-parameter tuning. This could be attributed to our network design and robust training procedure.   That said, in specific vertical applications, 
a more extensive tuning of hyper-parameters using any of the available methods~\cite{citeHyper} could be deployed for even larger gains.

The hyper-parameters and their default values in our network are:
the number of filters $p=32$ that controls the size of the first layer of the temporal transformer, 
the window size $w$ of the first convolution layer.  The hyper-parameter $w$ also determines the size of the context key used for attention.  If $w$ is very small, the context size may be inadequate, and if it is too large compared to the size of each series we may over-smooth patterns.  We use $10$ by default. When the average size of a missing block is large ($> 100$) we use $w=20$ to gather a larger context. 
The number of attention heads $n_{\mathrm{head}}$ is four and embedding size $d_i$ is taken to be $10$ in all our experiments. 
}

\section{Experiments}
We present results of our experiments on ten datasets under four different missing value scenarios. We compare imputation accuracy of several methods  spanning both traditional and deep learning and approaches in Sections~\ref{sec:expt:trad} and ~\ref{subsec:expt:point}. We then perform an ablation study to evaluate the various design choices of \sysname\ in Section~\ref{sec:expt:ablation}.  In Section~\ref{subsec:runtime} we compare different methods on running time. Finally, in Section~\ref{subsec:analysis} we 
highlight the importance of accurate imputation algorithms 
on downstream analytics.

\subsection{Experiment Setup}
\subsubsection{Datasets}
\begin{table}[]
    \centering
    \begin{tabular}{|l|r|r|l|l|}
        \hline
        Dataset & Number & Length & Repetitions & Relatedness \\ 
        & of TS  & of TS & within TS & across series \\ \hline \hline
        AirQ & 10 & 1k & Moderate & High\\ \hline
        Chlorine & 50 & 1k & High & High\\ \hline
        Gas & 100 & 1k & High & Moderate\\ \hline
        Climate & 10 & 5k & High & Low\\ \hline
        Electricity & 20 & 5k & High & Low\\ \hline
        Temperature & 50 & 5k & High & High\\ \hline
        Meteo & 10 & 10k & Low & Moderate\\ \hline
        BAFU & 10 & 50k & Low & Moderate\\ \hline
        JanataHack & 76*28 & 134 & Low & High\\ \hline
        M5 & 10*106 & 1941 & Low & Low\\ \hline
    \end{tabular}
    \caption{Datasets: All except the last two have one categorical dimension. Qualitative judgements on the repetitions of patterns along time and across series appear in the last two columns.}
    \label{tab:expt:datasets}
\end{table}
We experiments on eight datasets  used in earlier papers on missing value imputation~\cite{khayati2020mind}.  In addition, due to the lack of multi-dimensional datasets in previous works, we introduce two new datasets, ``JanataHack'', ``M5''. Table \ref{tab:expt:datasets} presents a summary along with qualitative judgements of their properties.

\noindent{\bf AirQ} brings air quality measurements collected from 36 monitoring stations in China from 2014 to 2015. AirQ time series contain both repeating patterns and jumps, and also strong correlations across time series. Replicating setup \cite{khayati2020mind}, we filter the dataset to get 10 time series of 1000 length.\\
\noindent{\bf Chlorine} simulates a drinking water distribution system on the concentration of chlorine in 166 junctions over 15 days in 5 minutes interval. This dataset contains clusters of similar time series which exhibit repeating trends.\\
\noindent{\bf Gas} shows gas concentration between 2007 and 2011 from a gas delivery platform of ChemoSignals Laboratory at UC San Diego. \\
\noindent{\bf Climate} is monthly climate data from 18 stations over 125 locations in North America between 1990 and 2002. These time series are irregular and contain sporadic spikes.

\noindent{\bf Electricity} is on household energy consumption collected every minute between 2006 and 2010 in France. \\
\noindent{\bf Temperature} contains temperature from climate stations in China from 1960 to 2012. These series are highly correlated.

\noindent{\bf MeteoSwiss} is weather
from different Swiss cities from 1980 to 2018 and contains repeating trends with sporadic anomalies.

\noindent{\bf BAFU} consists of water discharge data by the BundesAmt Für Umwelt (BAFU), collected from Swiss rivers from 1974 to 2015. These time series exhibit synchronized irregular trends.

\noindent{\bf JanataHack} is a multidimensional time series dataset\footnote{\url{https://www.kaggle.com/vin1234/janatahack-demand-forecasting-analytics-vidhya}} which consists of sales data spanning over 130 weeks, for 76 stores and 28 products (termed "SKU"). 

{\color{black} \noindent{\bf Walmart M5} made available by Walmart, involves the daily unit sales of $3049$ products sold in 10 stores in the USA spanning 5 years. 
Since most of the 3,049 items have 0 sales, we retain the 106 most selling items averaged over stores. This gives us a 2 dimensional data of sales of 106 items across 10 stores.}

\subsubsection{Missing Scenarios Description}
\label{subsec:missing_scenarios}
We experiment with four missing scenarios\cite{khayati2020mind} considered to be the common missing patterns encountered in real datasets. 
Here we consider continuous chunks of missing values termed as blocks. 
We also consider a scenario with point missing values scattered throughout the dataset in Sec \ref{sec:expt:ablation}.
\myparagraph{Missing Completely at Random (MCAR)} Each incomplete time series has 10\% of its data missing. The missing data is in randomly chosen blocks of constant size 10. We experiment with different \% of incomplete time series.

\myparagraph{Missing Disjoint (MissDisj)} Here we consider disjoint blocks to be missing. 
Block size is $T/N$, where $T$ is the length of time series, and $N$ is the number of time series. For $i$th time series the missing block ranges from time step $\frac{iT}{N}$ to $\frac{(i+1)T}{N}-1$, which ensures that 
missing blocks do not overlap across series.

\myparagraph{Missing Overlap (MissOver)} A slight modification on MissDisj, MissOver has block size of $2*T/N$ for all time series except the last one for which the block size is still $T/N$. For the $i$-th time series the missing block ranges from time step $\frac{iT}{N}$ to $\frac{(i+2)T}{N}-1$, which causes an overlap between missing blocks of series $i$ with $i-1$ and $i+1$ 
\myparagraph{Blackout} considers a scenario where all time series have missing values for the same time range. Given a block size $s$ all time series have values missing from $t$ to $t+s$, where $t$ is fixed to be $5\%$.
We vary the block size $s$ from 10 to 100.

\subsubsection{\bf Methods Compared}
We compare with methods from both conventional and deep learning literature. 

\noindent{\bf CDRec}\cite{khayati2019scalable} is one of the top performing recent Matrix Factorisation based technique which uses iterative Centroid Decomposition. \\ 
\noindent{\bf DynaMMO}\cite{li2009dynammo} is a probabilistic method that uses Kalman Filters to model co-evolution of subsets of similar time series. \\
\noindent{\bf TRMF}~\cite{yu2016temporal} is a matrix factorisation augmented with an auto-regressive temporal model.  \\
\noindent{\bf SVDImp}~\cite{troyanskaya2001missing} is a basic matrix factorisation based technique which imputes using top k vectors in SVD factorisation.\\
\noindent{\bf BRITS}~\cite{cao2018brits} is a recent Deep learning techniques that uses a Bidirectional RNN that takes as input all the series' values at time $t$. \\
{\color{black}
\noindent{\bf GPVAE}~\cite{yoon2018estimating} a deep learning method that uses Gaussian process in the low dimensional latent space representation of the data. GPVAE uses Variational Autoencoder to generate the imputed values in the original data space. \\
\noindent{\bf Transformer}~\cite{Vaswani2017} is a deep learning method that uses a multi-head self-attention based architecture to impute the missing values in time-series. 
}

\subsubsection{Other Experiment Details}
\myparagraph{Platforms} Our experiments are done on the Imputation Benchmark\footnote{\url{https://github.com/eXascaleInfolab/bench-vldb20}} for comparisons with conventional methods. The benchmark lacks in support for deep learning based algorithms hence we compare our numbers for those outside this framework. 

\myparagraph{Evaluation metric} We use Mean Absolute Error as our evaluation metric. 




\subsection{Visual Comparison of Imputation Quality}
\begin{figure}
\centering
\begin{tikzpicture}
\begin{groupplot}[group style={group size= 1 by 2,vertical sep=0cm,ylabels at=edge left},width=0.5\textwidth,height=0.20\textwidth]
\nextgroupplot[label style={font=\Large},
xtick=\empty,
xmin=0,
xmax=50,
ymin=-1.2,
ymax=1.5,
ytick style={draw=none},
legend style={at={($(0,0)+(1cm,1cm)$)},legend columns=4,fill=none,draw=black,anchor=center,align=center},
legend to name=fred,
mark size=1pt]
\addplot [black,mark=square*] coordinates {(1,-0.414889) (2,-0.568087) (3,-0.772350) (4,-0.925547) (5,-0.951080) (6,-0.976613) (7,-1.002150) (8,-0.925547) (9,-0.848949) (10,-0.363823)
};
\addplot [forget plot, black,mark=square*] coordinates {(11,0.555361) (12,0.631960) (13,0.376631) (14,0.223433) (15,0.070236) (16,0.019170) (17,0.044703) (18,0.019170) (19,0.351098) (20,0.810690)  
};
\addplot [forget plot, black,mark=square*] coordinates {(21,-0.354039) (22,-0.046337) (23,-0.148904) (24,-0.354039) (25,-0.354039) (26,-0.131810) (27,0.090420) (28,0.004947) (29,-0.097620) (30,0.620352)
};
\addplot [forget plot, black,mark=square*] coordinates {(31,-0.354039) (32,-0.747215) (33,-0.695931) (34,-0.456607) (35,-0.576269) (36,-0.542080) (37,-0.285661) (38,-0.080526) (39,0.381028) (40,0.483596)
};
\addplot [forget plot, black,mark=square*] coordinates {(41,1.475080) (42,1.167380) (43,0.979339) (44,0.329744) (45,0.312650) (46,0.534879) (47,0.945150) (48,0.551974) (49,0.483596) (50,0.517785)  
};
\addplot[red!70!black,mark=*]  coordinates {(1,0.377263) (2,-0.180590) (3,-0.130656) (4,-0.320763) (5,-0.545344) (6,-0.798426) (7,-0.824231) (8,-0.800857) (9,-0.891878) (10,-0.570796)
};
\addplot[forget plot, red!70!black,mark=*]  coordinates {(11,-0.018781) (12,0.153109) (13,-0.103472) (14,-0.247584) (15,-0.316140) (16,-0.887740) (17,-1.212370) (18,-0.874884) (19,0.104480) (20,0.362578) 
};
\addplot[forget plot, red!70!black,mark=*]  coordinates {(21,-0.140274) (22,-0.114944) (23,-0.257806) (24,-0.390871) (25,-0.287100) (26,-0.873556) (27,-0.659559) (28,-0.003037) (29,-0.187970) (30,0.212798) 
};
\addplot[forget plot, red!70!black,mark=*]  coordinates {(31,-0.084884) (32,-0.358025) (33,-0.397914) (34,-0.165218) (35,-0.285136) (36,-0.115910) (37,-0.088864) (38,0.650366) (39,1.324010) (40,1.412800)  
};
\addplot[forget plot, red!70!black,mark=*]  coordinates {(41,1.185970) (42,0.796645) (43,0.631419) (44,-0.294583) (45,-0.406536) (46,0.092161) (47,0.302551) (48,0.093460) (49,0.438395) (50,0.845026)  
};
\addplot[brown,mark=star] coordinates {(1,0.156504) (2,-0.087257) (3,-0.312328) (4,-0.453565) (5,-0.549715) (6,-0.651066) (7,-0.670796) (8,-0.661337) (9,-0.704424) (10,-0.525645)  
};

\addplot[forget plot, brown,mark=star] coordinates {(11,0.105762) (12,0.115198) (13,-0.043861) (14,-0.171330) (15,-0.148568) (16,-0.218026) (17,-0.251037) (18,-0.171568) (19,0.279208) (20,0.610292)  
};

\addplot[forget plot, brown,mark=star] coordinates {(21,-0.494509) (22,-0.321919) (23,-0.356131) (24,-0.483609) (25,-0.506485) (26,-0.533336) (27,-0.479690) (28,-0.331546) (29,0.017711) (30,0.462646)  
};

\addplot[forget plot, brown,mark=star] coordinates {(31,-0.283857) (32,-0.399765) (33,-0.464296) (34,-0.437765) (35,-0.362697) (36,-0.213132) (37,-0.014548) (38,0.469417) (39,0.844598) (40,0.990726)  
};

\addplot[forget plot, brown,mark=star] coordinates {(41,1.161680) (42,0.708668) (43,0.373171) (44,-0.236787) (45,-0.383092) (46,-0.208551) (47,0.031992) (48,0.219513) (49,0.452191) (50,0.392891)  
};

\addplot [blue,mark=diamond*] coordinates {(1,-0.078438) (2,-0.217998) (3,-0.580806) (4,-0.727121) (5,-0.865446) (6,-1.002165) (7,-0.978400) (8,-0.857621) (9,-0.800802) (10,-0.332229) 
};
\addplot [forget plot, blue,mark=diamond*] coordinates {(11,0.434284) (12,0.392550) (13,0.172372) (14,0.068014) (15,-0.018340) (16,-0.097801) (17,-0.081594) (18,-0.067918) (19,0.261959) (20,0.851163) 
};
\addplot [forget plot, blue,mark=diamond*] coordinates { (21,-0.281664) (22,-0.175742) (23,-0.179463) (24,-0.335860) (25,-0.437542) (26,-0.338008) (27,-0.225707) (28,-0.087221) (29,0.179038) (30,0.657779) 
};
\addplot [forget plot, blue,mark=diamond*] coordinates { (31,-0.257697) (32,-0.512213) (33,-0.658519) (34,-0.600982) (35,-0.645362) (36,-0.528507) (37,-0.431607) (38,0.080313) (39,0.397178) (40,0.675988) 
};
\addplot [forget plot, blue,mark=diamond*] coordinates {(41,1.527805) (42,1.168416) (43,0.972631) (44,0.494802) (45,0.516081) (46,0.699148) (47,0.743220) (48,0.633758) (49,0.573868) (50,0.505185)  
};
\addplot[black,mark=none] coordinates {(10, -2) (10, 4)};
\addplot[black,mark=none] coordinates {(20, -2) (20, 4)};
\addplot[black,mark=none] coordinates {(30, -2) (30, 4)};
\addplot[black,mark=none] coordinates {(40, -2) (40, 4)};

\addlegendentry{Ground Truth};
\addlegendentry{CDRec};
\addlegendentry{DynaMMO};
\addlegendentry{\sysname};          

\coordinate (c1) at (rel axis cs:0,1);

\nextgroupplot[label style={font=\Large},
tick label style={font=\large},
xmin=0,
xmax=50,
ymin=-1.5,
ymax=3.5,
xtick = {5,15,25,35,45},
ytick={-2,0,2},
xticklabels = {Block1,Block2,Block3,Block4,Block5},
xtick style={draw=none},
ytick style={draw=none},
mark size=1pt]
\addplot [black,mark=square*] coordinates {(1,1.413240) (2,0.906879) (3,-0.287616) (4,0.088909) (5,0.647206) (6,0.751075) (7,0.880911) (8,1.166550) (9,2.049440) (10,1.919600)  
};
\addplot [forget plot, black,mark=square*] coordinates {(11,-0.117444) (12,0.525530) (13,-0.032842) (14,-0.032842) (15,0.271724) (16,0.204043) (17,0.508610) (18,0.999301) (19,2.099130) (20,2.894380)   
};
\addplot [forget plot, black,mark=square*] coordinates {(21,-0.133840) (22,-0.293066) (23,-0.072600) (24,-0.550278) (25,-0.770745) (26,-0.672759) (27,-0.819737) (28,0.135619) (29,1.654390) (30,2.830210)  
};
\addplot [forget plot, black,mark=square*] coordinates { (31,0.197960) (32,0.063419) (33,-0.322266) (34,-0.044214) (35,-0.026275) (36,-0.089061) (37,0.215899) (38,0.745095) (39,2.234020) (40,2.969510)   
};
\addplot [forget plot, black,mark=square*] coordinates {(41,-0.687402) (42,-0.515071) (43,-0.554840) (44,-0.541584) (45,-0.833219) (46,-0.713914) (47,-0.660889) (48,0.200762) (49,1.897550) (50,2.348260)  
};
\addplot[red!70!black,mark=*]  coordinates {(1,1.849610) (2,1.871250) (3,1.892890) (4,1.914540) (5,1.936180) (6,1.957820) (7,1.979460) (8,2.001100) (9,2.022740) (10,2.044380) 
};
\addplot[forget plot, red!70!black,mark=*]  coordinates { (11,1.257060) (12,1.450560) (13,1.644050) (14,1.837550) (15,2.031050) (16,2.224540) (17,2.418040) (18,2.611540) (19,2.805040) (20,2.998530) 
};
\addplot[forget plot, red!70!black,mark=*]  coordinates {(21,0.401050) (22,0.695256) (23,0.989462) (24,1.283670) (25,1.577870) (26,1.872080) (27,2.166290) (28,2.460490) (29,2.754700) (30,3.048900) 
};
\addplot[forget plot, red!70!black,mark=*]  coordinates {(31,1.500660) (32,1.651490) (33,1.802330) (34,1.953160) (35,2.104000) (36,2.254840) (37,2.405670) (38,2.556510) (39,2.707350) (40,2.858180) 
};
\addplot[forget plot, red!70!black,mark=*]  coordinates {(41,0.254779) (42,0.547164) (43,0.839549) (44,1.131930) (45,1.424320) (46,1.716700) (47,2.009090) (48,2.301470) (49,2.593860) (50,2.886240)  
};

\addplot[brown,mark=star] coordinates {(1,1.531860) (2,1.581630) (3,1.644250) (4,1.719640) (5,1.805320) (6,1.895940) (7,1.983090) (8,2.055360) (9,2.098890) (10,2.098300) };
\addplot[forget plot, brown,mark=star] coordinates {(11,0.913840) (12,1.028080) (13,1.162920) (14,1.323570) (15,1.512970) (16,1.730860) (17,1.972990) (18,2.230620) (19,2.490600) (20,2.736010)};
\addplot[forget plot, brown,mark=star] coordinates { (21,0.240437) (22,0.378191) (23,0.537924) (24,0.727981) (25,0.955180) (26,1.223700) (27,1.534000) (28,1.881950) (29,2.258470) (30,2.649800)};
\addplot[forget plot, brown,mark=star] coordinates {(31,1.144350) (32,1.249360) (33,1.374360) (34,1.523500) (35,1.698460) (36,1.897600) (37,2.115260) (38,2.341360) (39,2.561630) (40,2.758360)};
\addplot[forget plot, brown,mark=star] coordinates {(41,0.321312) (42,0.453297) (43,0.606865) (44,0.790670) (45,1.012130) (46,1.276300) (47,1.584810) (48,1.934910) (49,2.319030) (50,2.724870)};

\addplot [blue,mark=diamond*] coordinates {(1,0.873680) (2,0.586259) (3,0.328305) (4,0.263862) (5,0.396760) (6,0.597257) (7,0.840426) (8,1.297004) (9,1.758366) (10,1.895521)};
\addplot [forget plot, blue,mark=diamond*] coordinates {(11,0.665421) (12,0.846143) (13,0.813302) (14,0.467279) (15,0.209087) (16,0.181149) (17,0.352080) (18,0.927669) (19,1.794154) (20,2.378234) };
\addplot [forget plot, blue,mark=diamond*] coordinates {(21,0.177819) (22,0.390226) (23,0.362192) (24,0.029787) (25,-0.135682) (26,-0.198076) (27,-0.205636) (28,0.413437) (29,1.591336) (30,2.396279)};
\addplot [forget plot, blue,mark=diamond*] coordinates { (31,0.929456) (32,0.645202) (33,0.403300) (34,0.412816) (35,0.533824) (36,0.513319) (37,0.493715) (38,0.976982) (39,2.031047) (40,2.887430) };
\addplot [forget plot, blue,mark=diamond*] coordinates { (41,-0.219655) (42,-0.062898) (43,-0.113565) (44,-0.412571) (45,-0.515318) (46,-0.558704) (47,-0.483313) (48,0.495419) (49,1.998899) (50,2.753356)};

\addplot[black,mark=none] coordinates {(10, -2) (10, 4)};
\addplot[black,mark=none] coordinates {(20, -2) (20, 4)};
\addplot[black,mark=none] coordinates {(30, -2) (30, 4)};
\addplot[black,mark=none] coordinates {(40, -2) (40, 4)};
\coordinate (c2) at (rel axis cs:1,1);

\end{groupplot}
\coordinate (c3) at ($(c1)!.5!(c2)$);
\node[below] at (c3 |- current bounding box.south)
{\pgfplotslegendfromname{fred}};

\end{tikzpicture}
\caption{Visualised Imputations on Electricity Dataset. The top row shows MCAR missing blocks while the bottom rows is for Blackout scenario.}
\label{fig:visualise}
\end{figure}
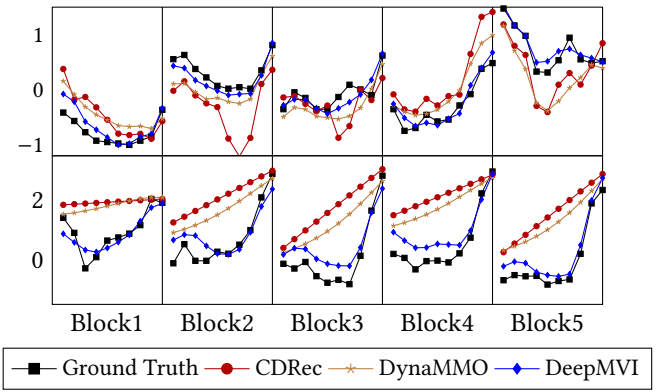

We start with a visual illustration of how \sysname's imputations compare with those of two of the best performing existing methods: CDRec and DynaMMO.  In Fig.~\ref{fig:visualise}, we visualize the imputations for different missing blocks on the Electricity dataset. First row is for MCAR scenario whereas second row is for Blackout scenario. 
First observe how \sysname (Blue) correctly captures both the shape and scale of actual values (Black) over a range of missing blocks. 
On the MCAR scenario CDRec gets the shape right, only in the first and fourth blocks, however it is off with scale. 
In the Blackout scenario, CDRec only linearly interpolates the values in missing block, whereas DynaMMO is only slightly inclined towards ground-truth. However, both CDRec and DynaMMO miss the trend during Blackout whereas \sysname\ successfully captures it because of careful pattern match within a series.

\subsection{Comparison on Imputation Accuracy}
\label{sec:expt:trad}
\begin{figure*}
\centering
\begin{tikzpicture}
\begin{groupplot}[group style={group size= 2 by 2},width=0.5\textwidth,height=0.20\textwidth]
\nextgroupplot[ybar=1pt, 
symbolic x coords={Chlorine,Temp,Gas,Meteo,BAFU},
xtick={Chlorine,Temp,Gas,Meteo,BAFU},
scaled y ticks = false,
legend pos = north west, 
ymin=0,
tickwidth=0pt,
ymajorgrids=true,
legend style={at={($(0,0)+(1cm,1cm)$)},legend columns=5,fill=none,draw=black,anchor=center,align=center},
title=MCAR,
legend to name=fred,
bar width=4pt]
\addplot+[red, fill=red!30!white] coordinates {(Chlorine, 0.067) (Temp, 0.126) (Gas, 0.181) (Meteo, 0.202) (BAFU, 0.203) };
\addplot+[brown!40!black,fill=brown!50!white] coordinates {(Chlorine, 0.057) (Temp, 0.126) (Gas, 0.204) (Meteo, 0.192) (BAFU, 0.184) };
\addplot+[cyan!40!black, fill=cyan!30!white] coordinates {(Chlorine, 0.063) (Temp, 0.126) (Gas, 0.204) (Meteo, 0.200) (BAFU, 0.216) };
\addplot+[gray, fill=black!40!white] coordinates {(Chlorine, 0.064) (Temp, 0.126) (Gas, 0.205) (Meteo, 0.200) (BAFU, 0.217) };
\addplot+[blue, fill=blue!40!white] coordinates {(Chlorine, 0.035) (Temp, 0.075) (Gas, 0.074) (Meteo, 0.096) (BAFU, 0.147) };

\legend{CDRec,DynaMMO,TRMF,SVDImp,\sysname};
\coordinate (c1) at (rel axis cs:0,1);

\nextgroupplot[ybar=1pt, 
symbolic x coords={Chlorine,Temp,Gas,Meteo,BAFU},
legend pos = north west, 
ymin=0,
title = MissDisj,
tickwidth=0pt,
ymajorgrids=true,
bar width=4pt]
\addplot+[red, fill=red!30!white] coordinates {(Chlorine, 0.071) (Temp, 0.129) (Gas, 0.180) (Meteo, 0.235) (BAFU, 0.177) };
\addplot+[brown!40!black,fill=brown!50!white] coordinates {(Chlorine, 0.070) (Temp, 0.128) (Gas, 0.223) (Meteo, 0.242) (BAFU, 0.209) };
\addplot+[cyan!40!black, fill=cyan!30!white] coordinates {(Chlorine, 0.073) (Temp, 0.128) (Gas, 0.225) (Meteo, 0.246) (BAFU, 0.202) };
\addplot+[gray, fill=black!40!white]  coordinates {(Chlorine, 0.067) (Temp, 0.128) (Gas, 0.226) (Meteo, 0.247) (BAFU, 0.203) };
\addplot+[blue, fill=blue!40!white] coordinates {(Chlorine, 0.084) (Temp, 0.073) (Gas, 0.053) (Meteo, 0.091) (BAFU, 0.104) };

\coordinate (c2) at (rel axis cs:1,1);
\nextgroupplot[ybar=1pt, 
symbolic x coords={Chlorine,Temp,Gas,Meteo,BAFU},
legend pos = north west, 
ymin=0,
title=MissOver,
tickwidth=0pt,
ymajorgrids=true,
bar width=4pt]
\addplot+[red, fill=red!30!white] coordinates {(Chlorine, 0.108) (Temp, 0.127) (Gas, 0.177) (Meteo, 0.263) (BAFU, 0.175) };
\addplot+[brown!40!black,fill=brown!50!white] coordinates {(Chlorine, 0.102) (Temp, 0.127) (Gas, 0.216) (Meteo, 0.268) (BAFU, 0.191) };
\addplot+[cyan!40!black, fill=cyan!30!white] coordinates {(Chlorine, 0.116) (Temp, 0.128) (Gas, 0.217) (Meteo, 0.272) (BAFU, 0.188) };
\addplot+[gray, fill=black!40!white]  coordinates {(Chlorine, 0.106) (Temp, 0.128) (Gas, 0.218) (Meteo, 0.273) (BAFU, 0.188) };
\addplot+[blue, fill=blue!40!white] coordinates {(Chlorine, 0.107) (Temp, 0.079) (Gas, 0.069) (Meteo, 0.247) (BAFU, 0.155) };

\nextgroupplot[ybar=1pt, 
symbolic x coords={Chlorine,Temp,Gas,Meteo,BAFU},
legend pos = north west, 
ymin=0,
tickwidth=0pt,
ymajorgrids=true,
ymax = 1,
title=Blackout,
bar width=4pt]
\addplot+[red, fill=red!30!white] coordinates {(Chlorine, 0.090) (Temp, 0.098) (Gas, 0.754) (Meteo, 0.426) (BAFU, 0.045) };
\addplot+[brown!40!black,fill=brown!50!white] coordinates {(Chlorine, 0.070) (Temp, 0.201) (Gas, 0.774) (Meteo, 0.376) (BAFU, 0.054) };
\addplot+[cyan!40!black, fill=cyan!30!white] coordinates {(Chlorine, 0.640) (Temp, 1.241) (Gas, 1.030) (Meteo, 4.170) (BAFU, 6.099) };
\addplot+[gray, fill=black!40!white]  coordinates {(Chlorine, 0.092) (Temp, 0.818) (Gas, 0.748) (Meteo, 1.301) (BAFU, 0.835) };
\addplot+[blue, fill=blue!40!white] coordinates {(Chlorine, 0.035) (Temp, 0.123) (Gas, 0.154) (Meteo, 0.349) (BAFU, 0.117) };

\coordinate (c4) at (rel axis cs:0,0);

\end{groupplot}
\path (c1-|current bounding box.west)-- node[anchor=south,rotate=90] {\large Mean Absolute Error} (c4-|current bounding box.west);
\coordinate (c3) at ($(c1)!.5!(c2)$);
\node[below] at (c3 |- current bounding box.south)
{\pgfplotslegendfromname{fred}};

\end{tikzpicture}
\caption{Mean Absolute Errors (y-axis) on five other datasets (on x-axis) on all four scenarios -- MCAR, MissDisj, MissOver, and Blackout. Here, a fixed $x=10\%$ of the series in each dataset has missing blocks.}
\label{fig:main_plots_bar}
\end{figure*}
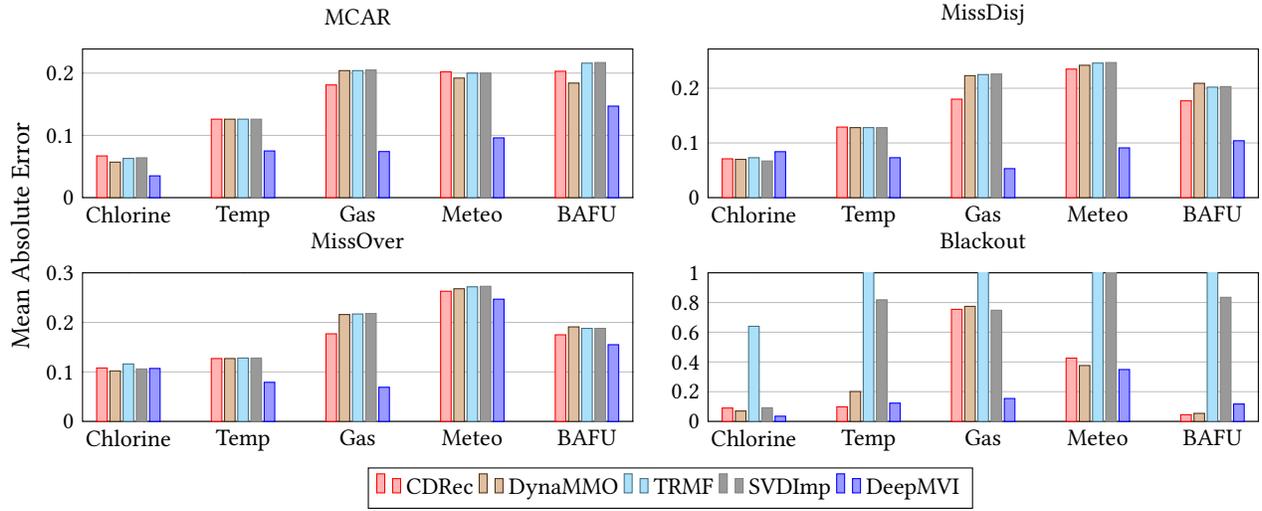
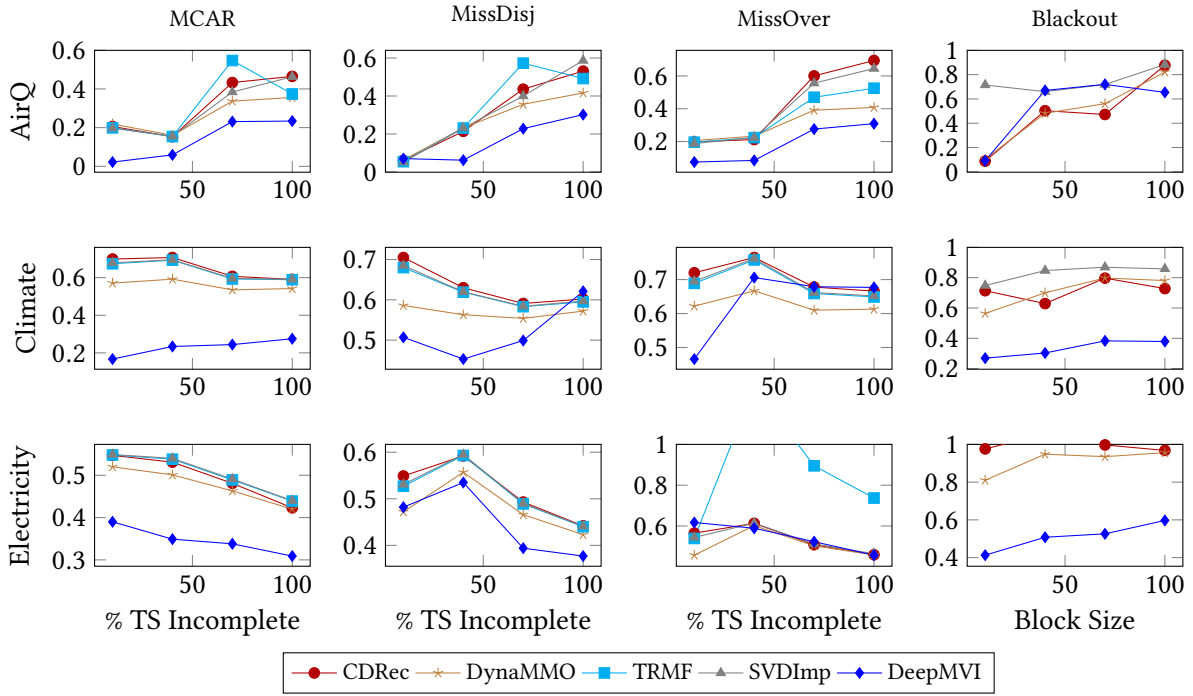
\begin{figure*}
\centering
\begin{tikzpicture}
\begin{groupplot}[group style={group size= 4 by 3,ylabels at=edge left},width=0.25\textwidth,height=0.18\textwidth]
\nextgroupplot[label style={font=\Large},
tick label style={font=\Large},
legend style={at={($(0,0)+(1cm,1cm)$)},legend columns=6,fill=none,draw=black,anchor=center,align=center},
ylabel = {AirQ},
legend to name=fred,
title = MCAR,
mark size=2pt]
\addplot[red!70!black,mark=*] coordinates {(10, 0.206) (40, 0.153) (70, 0.433) (100, 0.465) };
\addplot[brown,mark=star] coordinates {(10, 0.219) (40, 0.160) (70, 0.337) (100, 0.356) };
\addplot[cyan,mark=square*] coordinates {(10, 0.199) (40, 0.154) (70, 0.547) (100, 0.374) };
\addplot[gray,mark=triangle*] coordinates {(10, 0.197) (40, 0.155) (70, 0.384) (100, 0.463) };
\addplot[blue,mark=diamond*] coordinates {(10, 0.022) (40, 0.059) (70, 0.231) (100, 0.234) };
\addlegendentry{CDRec};    
\addlegendentry{DynaMMO};    
\addlegendentry{TRMF};          
\addlegendentry{SVDImp};          
\addlegendentry{\sysname};          
\coordinate (c1) at (rel axis cs:0,1);

\nextgroupplot[label style={font=\Large},
tick label style={font=\Large},
title = MissDisj,
mark size=2pt]
\addplot [red!70!black,mark=*] coordinates {(10, 0.061) (40, 0.214) (70, 0.436) (100, 0.531) };
\addplot [brown,mark=star] coordinates {(10, 0.063) (40, 0.232) (70, 0.356) (100, 0.416) };
\addplot [cyan,mark=square*] coordinates {(10, 0.054) (40, 0.231) (70, 0.573) (100, 0.492) };
\addplot [gray,mark=triangle*] coordinates {(10, 0.052) (40, 0.231) (70, 0.400) (100, 0.587) };
\addplot [blue,mark=diamond*] coordinates {(10, 0.070) (40, 0.062) (70, 0.228) (100, 0.302) };

\nextgroupplot[label style={font=\Large},
tick label style={font=\Large},
title = MissOver,
mark size=2pt]
\addplot [red!70!black,mark=*] coordinates {(10, 0.198) (40, 0.212) (70, 0.600) (100, 0.694) };
\addplot [brown,mark=star] coordinates {(10, 0.206) (40, 0.233) (70, 0.391) (100, 0.408) };
\addplot [cyan,mark=square*] coordinates {(10, 0.196) (40, 0.224) (70, 0.470) (100, 0.525) };
\addplot [gray,mark=triangle*] coordinates {(10, 0.188) (40, 0.224) (70, 0.556) (100, 0.645) };
\addplot [blue,mark=diamond*] coordinates {(10, 0.075) (40, 0.085) (70, 0.276) (100, 0.309) };

\nextgroupplot[label style={font=\Large},
tick label style={font=\Large},
title = Blackout,
ymax=1,
mark size=2pt]
\addplot [red!70!black,mark=*] coordinates {(10, 0.088) (40, 0.503) (70, 0.472) (100, 0.877) };
\addplot [brown,mark=star] coordinates {(10, 0.104) (40, 0.482) (70, 0.561) (100, 0.821) };
\addplot [cyan,mark=square*] coordinates {(10, 2.847) (40, 2.515) (70, 2.186) (100, 1.968) };
\addplot [gray,mark=triangle*] coordinates {(10, 0.715) (40, 0.660) (70, 0.719) (100, 0.883) };
\addplot [blue,mark=diamond*] coordinates {(10, 0.092) (40, 0.670) (70, 0.720) (100, 0.654) };

\coordinate (c2) at (rel axis cs:1,1);
\nextgroupplot[label style={font=\Large},
ylabel = {Climate},
tick label style={font=\Large},
mark size=2pt]
\addplot [red!70!black,mark=*] coordinates {(10, 0.699) (40, 0.707) (70, 0.607) (100, 0.591) };
\addplot [brown,mark=star] coordinates {(10, 0.571) (40, 0.592) (70, 0.535) (100, 0.542) };
\addplot [cyan,mark=square*] coordinates {(10, 0.674) (40, 0.693) (70, 0.593) (100, 0.589) };
\addplot [gray,mark=triangle*] coordinates {(10, 0.679) (40, 0.695) (70, 0.596) (100, 0.591) };
\addplot [blue,mark=diamond*] coordinates {(10, 0.167) (40, 0.234) (70, 0.244) (100, 0.275) };
\nextgroupplot[label style={font=\Large},
tick label style={font=\Large},
mark size=2pt]
\addplot [red!70!black,mark=*] coordinates {(10, 0.705) (40, 0.630) (70, 0.591) (100, 0.602) };
\addplot [brown,mark=star] coordinates {(10, 0.586) (40, 0.563) (70, 0.554) (100, 0.572) };
\addplot [cyan,mark=square*] coordinates {(10, 0.680) (40, 0.619) (70, 0.583) (100, 0.595) };
\addplot [gray,mark=triangle*] coordinates {(10, 0.685) (40, 0.620) (70, 0.584) (100, 0.597) };
\addplot [blue,mark=diamond*] coordinates {(10, 0.507) (40, 0.453) (70, 0.499) (100, 0.621) };

\nextgroupplot[label style={font=\Large},
tick label style={font=\Large},
mark size=2pt]
\addplot [red!70!black,mark=*] coordinates {(10, 0.720) (40, 0.765) (70, 0.678) (100, 0.666) };
\addplot [brown,mark=star] coordinates {(10, 0.622) (40, 0.667) (70, 0.610) (100, 0.613) };
\addplot [cyan,mark=square*] coordinates {(10, 0.689) (40, 0.758) (70, 0.659) (100, 0.649) };
\addplot [gray,mark=triangle*] coordinates {(10, 0.695) (40, 0.762) (70, 0.662) (100, 0.652) };
\addplot [blue,mark=diamond*] coordinates {(10, 0.466) (40, 0.706) (70, 0.679) (100, 0.677) };

\nextgroupplot[label style={font=\Large},
tick label style={font=\Large},
ymax=1,
mark size=2pt]
\addplot [red!70!black,mark=*] coordinates {(10, 0.714) (40, 0.629) (70, 0.797) (100, 0.728) };
\addplot [brown,mark=star] coordinates {(10, 0.564) (40, 0.700) (70, 0.798) (100, 0.781) };
\addplot [cyan,mark=square*] coordinates {(10, 2.941) (40, 3.144) (70, 3.181) (100, 3.284) };
\addplot [gray,mark=triangle*] coordinates {(10, 0.748) (40, 0.847) (70, 0.869) (100, 0.858) };
\addplot [blue,mark=diamond*] coordinates {(10, 0.269) (40, 0.303) (70, 0.383) (100, 0.379) };

\nextgroupplot[label style={font=\Large},
ylabel = {Electricity},
tick label style={font=\Large},
mark size=2pt,
xlabel = \% TS Incomplete]
\addplot [red!70!black,mark=*] coordinates {(10, 0.547) (40, 0.531) (70, 0.481) (100, 0.423) };
\addplot [brown,mark=star] coordinates {(10, 0.520) (40, 0.501) (70, 0.463) (100, 0.420) };
\addplot [cyan,mark=square*] coordinates {(10, 0.548) (40, 0.538) (70, 0.489) (100, 0.439) };
\addplot [gray,mark=triangle*] coordinates {(10, 0.549) (40, 0.540) (70, 0.491) (100, 0.440) };
\addplot [blue,mark=diamond*] coordinates {(10, 0.390) (40, 0.349) (70, 0.338) (100, 0.309) };
\nextgroupplot[label style={font=\Large},
tick label style={font=\Large},
mark size=2pt,
xlabel = \% TS Incomplete]
\addplot [red!70!black,mark=*] coordinates {(10, 0.549) (40, 0.592) (70, 0.493) (100, 0.442) };
\addplot [brown,mark=star] coordinates {(10, 0.472) (40, 0.557) (70, 0.466) (100, 0.423) };
\addplot [cyan,mark=square*] coordinates {(10, 0.527) (40, 0.593) (70, 0.489) (100, 0.440) };
\addplot [gray,mark=triangle*] coordinates {(10, 0.531) (40, 0.595) (70, 0.490) (100, 0.441) };
\addplot [blue,mark=diamond*] coordinates {(10, 0.482) (40, 0.535) (70, 0.394) (100, 0.377) };

\nextgroupplot[label style={font=\Large},
tick label style={font=\Large},
ymax=1,
mark size=2pt,
xlabel = \% TS Incomplete]
\addplot [red!70!black,mark=*] coordinates {(10, 0.565) (40, 0.613) (70, 0.508) (100, 0.459) };
\addplot [brown,mark=star] coordinates {(10, 0.457) (40, 0.602) (70, 0.503) (100, 0.458) };
\addplot [cyan,mark=square*] coordinates {(10, 0.539) (40, 1.284) (70, 0.895) (100, 0.737) };
\addplot [gray,mark=triangle*] coordinates {(10, 0.544) (40, 0.614) (70, 0.512) (100, 0.464) };
\addplot [blue,mark=diamond*] coordinates {(10, 0.617) (40, 0.589) (70, 0.523) (100, 0.457) };

\nextgroupplot[label style={font=\Large},
tick label style={font=\Large},
ymax=1,
mark size=2pt,
xlabel = Block Size]
\addplot [red!70!black,mark=*] coordinates {(10, 0.976) (40, 1.055) (70, 0.997) (100, 0.968) };
\addplot [brown,mark=star] coordinates {(10, 0.810) (40, 0.948) (70, 0.935) (100, 0.956) };
\addplot [cyan,mark=square*] coordinates {(10, 1.978) (40, 2.146) (70, 2.221) (100, 2.264) };
\addplot [gray,mark=triangle*] coordinates {(10, 1.208) (40, 1.124) (70, 1.051) (100, 1.058) };
\addplot [blue,mark=diamond*] coordinates {(10, 0.413) (40, 0.508) (70, 0.526) (100, 0.597) };

\end{groupplot}
\coordinate (c3) at ($(c1)!.5!(c2)$);
\node[below] at (c3 |- current bounding box.south)
{\pgfplotslegendfromname{fred}};

\end{tikzpicture}
\caption{Mean Absolute Errors (y-axis) on three datasets along  rows and under four missing scenarios along columns. X-axis is percent of time-series with a missing block for MCAR, MissDisj, MissOver and size of the missing block for Blackout.}
\label{fig:graph}
\end{figure*}


Given the large number of datasets, methods, missing scenarios and missing sizes we present our numbers in stages. First in  Figure~\ref{fig:main_plots_bar} we show comparisons in MAE of all conventional methods on five datasets under a fixed $x=$10\% of series with missing values in MCAR, MissDisj, MissOver and all series in Blackout with a block size of 10.  Then, in Figure~\ref{fig:graph} we show more detailed MAE numbers on three datasets (AirQ, Climate and Electricity)  where we vary the percent of series with missing values ($x$) from 10 to 100 for MCAR, MissDisj, MissOver and the block size from 10 to 100 in Blackout.  From these comparisons across eight datasets we make the following observations:

First, observe that \sysname\ is better or comparable to all other methods under all missing values scenarios and all datasets.  
Our gains are particularly high in the Blackout scenario seen in the last column in graphs in Figure~\ref{fig:graph} and in the bottom-right graph of Figure~\ref{fig:barplots}.  For accurate imputation in Blackouts, we need to exploit signals from other locations of the same series. Matrix factorisation based methods such as SVDImp and TRMF fail in doing so and rely heavily on correlation across time series. TRMF's temporal regularisation does not seem to be helping in capturing long term temporal correlations.  DynaMMO and CDRec 
are able to capture within time series dependencies better than matrix factorisation methods. But they are still much worse than \sysname, particularly on  Gas in Figure~\ref{fig:barplots}, and Climate, Electricity in Figure~\ref{fig:graph}. 

In the MissDisj/MissOver scenario where the same time range is not missing across all time series, methods that effectively exploit relatedness across series perform better on datasets with highly correlated series such as Chlorine and Temp. 
Even in these scenarios we provide as much as 50\% error reduction compared to existing methods.

MCAR is the most interesting scenario for our analysis. Most of the baselines are geared towards capturing either inter or intra TS correlation but none of them are able to effectively combine and exploit both. MCAR owing to small block size and random missing position can benefit from both inter and intra correlation which are fully exploited by our model. \sysname\ achieves strictly better numbers than all the baselines on all the datasets. For Climate and Electricity datasets, 
we reduce errors between 20\% and  70\% as seen in the first column of Figure~\ref{fig:graph}. 






\begin{table*}[]
    \centering
    \begin{tabular}{|l|c|c||c|c|c|c|c|c|}
        \hline
        \multirow{2}{*}{Model} & Walmart M5 & JantaHack & \multicolumn{2}{|c|}{Climate} & \multicolumn{2}{|c|}{Electricity} & \multicolumn{2}{|c|}{Meteo} \\ 
        \cline{2-9}
          & MCAR & MCAR & MCAR & Blackout & MCAR & Blackout & MCAR & Blackout \\ \hline \hline
        
        BRITS \cite{cao2018brits} & 0.69 & 0.22 & 0.26 & 0.69 & 0.28 & 1.16 & 0.19 & 0.77\\ \hline
        GPVAE\cite{fortuin2020gp} & 0.60 & 0.28 & 0.43 & 0.81  & 0.33 & 1.08 & 0.26 & 1.31\\ \hline
        Transformer \cite{Vaswani2017} & 0.56 & 0.24 & 0.29 & 0.67  & 0.36 & 0.97 & 0.29 & 0.48\\ \hline
        \sysname (Ours)& 0.53 & 0.16 & 0.28 & 0.38  & 0.31 & 0.60 & 0.20 & 0.46\\ \hline
    \end{tabular}
    \caption{Comparison with Deep Learning Methods. Blackout has missing blocks of size 100, and MCAR has missing values in 100\% of the time series.}
    \label{tab:expt:dl}
\end{table*}
\begin{figure*}[ht]
\centering
\begin{tikzpicture}
\begin{groupplot}[group style={group size= 3 by 1,ylabels at=edge left},width=0.30\textwidth,height=0.25\textwidth]
\nextgroupplot[label style={font=\Large},title=AirQ,
tick label style={font=\Large},
legend style={at={($(0,0)+(1cm,1cm)$)},legend columns=1,fill=none,draw=black,anchor=center,align=left,legend cell align=left,font=\small},
ylabel = {MAE},
legend to name=fred,
mark size=2pt]
\addplot [red,mark=square*] coordinates {(10, 0.036) (40, 0.08) (70, 0.24) (100, 0.25) };
\addplot [green,mark=*] coordinates {(10, 0.03) (40, 0.07) (70, 0.23) (100, 0.24) };
\addplot [brown,mark=otimes*] coordinates {(10, 0.253) (40, 0.231) (70, 0.394) (100, 0.318) };
\addplot [blue,mark=diamond*] coordinates {(10, 0.022) (40, 0.059) (70, 0.231) (100, 0.234) };

\addlegendentry{No Temporal Transformer};    
\addlegendentry{No Context Window};    
\addlegendentry{No Kernel Regression};          
\addlegendentry{\sysname};          
\coordinate (c1) at (rel axis cs:0,1);

\nextgroupplot[label style={font=\Large},title=Climate,
tick label style={font=\Large},
mark size=2pt]
\addplot [red,mark=square*] coordinates {(10, 0.546) (40, 0.613) (70, 0.597) (100, 0.609) };
\addplot [green,mark=*] coordinates {(10, 0.202) (40, 0.354) (70, 0.345) (100, 0.369) };
\addplot [brown,mark=otimes*]  coordinates {(10, 0.155) (40, 0.332) (70, 0.310) (100, 0.336) };
\addplot[blue,mark=diamond*]  coordinates {(10, 0.167) (40, 0.284) (70, 0.298) (100, 0.340) };
\nextgroupplot[label style={font=\Large},title=Electricity,
tick label style={font=\Large},
mark size=2pt]
\addplot [red,mark=square*]  coordinates {(10, 0.589) (40, 0.515) (70, 0.464) (100, 0.410) };
\addplot [green,mark=*] coordinates {(10, 0.432) (40, 0.424) (70, 0.418) (100, 0.359) };
\addplot  [brown,mark=otimes*] coordinates {(10, 0.392) (40, 0.344) (70, 0.343) (100, 0.322) };
\addplot[blue,mark=diamond*]  coordinates {(10, 0.390) (40, 0.349) (70, 0.338) (100, 0.309) };
\coordinate (c2) at (rel axis cs:1,1);

\end{groupplot}
\coordinate (c3) at ($(c2)$);
\node[right] at (c3 |- current bounding box.west)
{\pgfplotslegendfromname{fred}};

\end{tikzpicture}
\caption{Ablation Study done via 3 datasets represented by different plots AirQ,Climate,Electricity on MCAR scenario. y-axis shows MAE and x-axis \% of Missing TS.}
\label{fig:ablation}
\end{figure*}
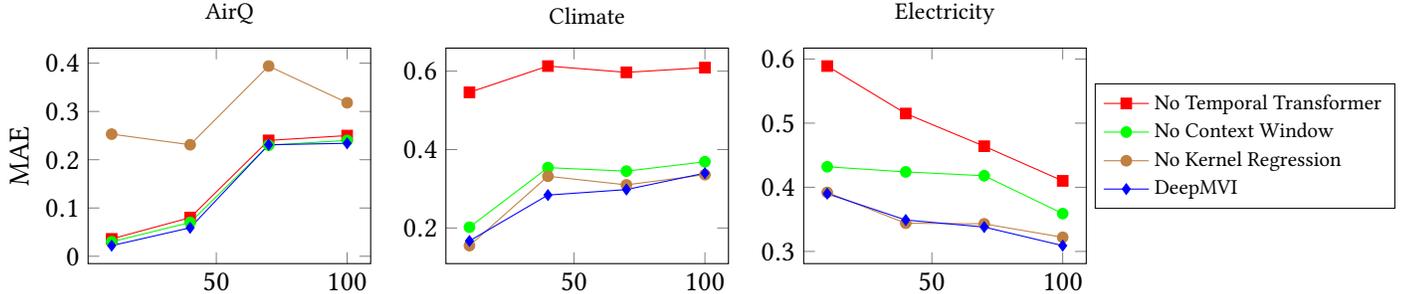

\subsection{Comparison with Deep Learning Methods}
\label{subsec:expt:point}
{\color{black} 
We compare our method with with two state-of-the-art deep learning imputation models, along with a vanilla transformer model. We use the official implementation of BRITS and GP-VAE to report these numbers. We present MAE numbers in Table \ref{tab:expt:dl}. 
First consider the comparison on the two multi-dimensional datasets: M5 and JanataHack. Both have store and items as the two dimensions in addition to time (Table \ref{tab:expt:datasets}).
We experiment  in the MCAR scenario with $x=100\%$ time-series with a missing block.  We find that \sysname\  outperforms all the other imputation models on both these datasets. The decrease in MAE is especially significant for JantaHack which has high correlations across different stores for given products. 

We next present our numbers on Climate, Electricity and Meteo, on MCAR and Blackout. Here too \sysname\ is either the best or close to the best in all dataset-scenario combinations.  On the Blackout scenario our method is significantly better than BRITS, the state-of-the-art. We attribute this to our method of creating artificial blackouts around training indices. In contrast, the BRITS model depends too much on immediate temporal neighborhood during training.
We see that the Transformer model can capture periodic correlations within time series such as those in Climate MCAR. However it fails to capture more subtle non-periodic repeating pattern which requires attention on window feature vectors. Such patterns are prevalent in Electricity and Meteo datasets. 
}

\subsection{Justifying Design Choices of \sysname}
\label{sec:expt:ablation}
\sysname\ introduces a Temporal transformer with an innovative left-right window feature to capture coarse-grained context, a fine-grained local signal, and a kernel regression module that handles multi-dimensional data. Here we perform an ablation study to dissect the role of each of these parts.

 


\subsubsection{Role of Context Window Features}
We study the role of query and key used in our Temporal Transformer module.
Our query/key consists of concatenated window features of previous and next block arithmetically added with positional encoding. Positional encoding encode the relative positions and have no information pertaining to the context of the block where imputation needs to be performed. A question to be asked here was whether the contextual information around a missing block help in a better attention mechanism or whether the attention mechanism just ignores this contextual information doing a fixed periodic imputation. Figure~\ref{fig:ablation} shows this method (Green). These experiments are on MCAR and x-axis is increasing \% of missing TS.
Comparing the green and blue, 
we see that our window context features did help on two of the three datasets, with the impact on Electricity being quite significant. 
This might be attributed to the periodic nature of the climate dataset compared to non-periodic but strongly contextual information in electricity. 

\subsubsection{Role of Temporal Transformer and Kernel Regression}
In Figure~\ref{fig:ablation} we present error without the Temporal Transformer Module(Red) and without Kernel Regression Module(Brown).   We see some interesting trends here. On Climate and Electricity where each series is large (5k) with repeated patterns across series, we see that dropping the temporal transformer causes large jumps in error. On Climate error jumps from 0.15 to 0.55 with 10\% missing!  In AirQ we see little impact. However, on this data dropping Kernel regression causes a large increase in error jumping from 0.04 to 0.25 on 10\% missing.  Kernel regression does not help much beyond temporal transformer on Climate and Electricity.  These experiments show that \sysname\ is capable of   
combining both factors and determining the dominating correlation via the training process. 
\subsubsection{Role of Fine-Grained Local Signal}
\label{subsubsec:expts_fine_grained}
\begin{figure}
\centering
\begin{tikzpicture}
\begin{groupplot}[group style={group size= 1 by 1,ylabels at=edge left},width=0.50\textwidth,height=0.25\textwidth]
\nextgroupplot[label style={font=\Large},
tick label style={font=\Large},
ylabel = {MAE},
legend style={at={(1,0.5)},anchor= east,legend columns=1,legend cell align=left,font=\small},
mark size=2pt]
\addplot[red!70!black,mark=*] coordinates {(1, 0.440) (2, 0.445) (5, 0.436) (10,0.423)};
\addplot [blue,mark=diamond*,dashed] coordinates {(1, 0.382) (2, 0.367) (5, 0.311) (10,0.309)};
\addplot [blue,mark=diamond*] coordinates {(1, 0.300) (2, 0.311) (5, 0.309) (10,0.309)};
\addlegendentry{CDRec};  
\addlegendentry{No FineGrained};          
\addlegendentry{FineGrained};          
\end{groupplot}

\end{tikzpicture}
\caption{MAE (y-axis) vs missing block size (x-axis) with 10\% missing on Climate Dataset. Gains from fine-grained local signal decreases with increasing block size.}
\label{fig:finegrained}
\end{figure}
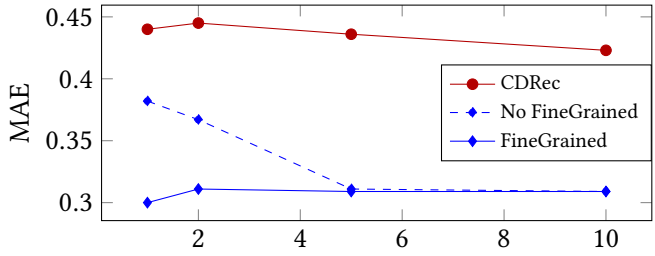

(Equation~\ref{eqn:finegrained}). This signal is most useful for small missing blocks. Hence we modify the MCAR missing scenario such that missing percentage from all time series is still 10\%, but the missing block size is varied from 1 to 10.  Figure~\ref{fig:finegrained} shows the results where we %
%
compare our MAE with and without fine grained local signal with CDRec algorithm on the Climate Dataset.
The plot shows that including fine grained signal helps improve accuracy over a model which ignores the local information. Also the gain in accuracy with fine grained local signal diminishes with increasing block size which is to be expected.




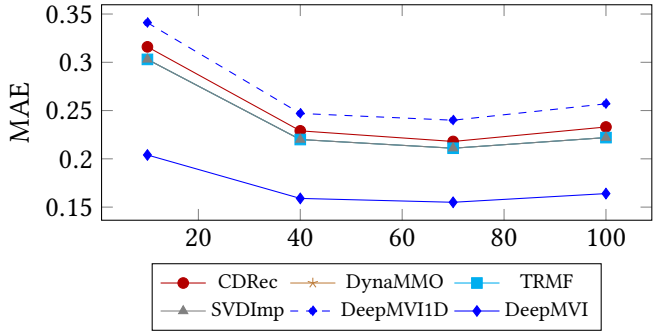
\begin{figure}
\centering
\begin{tikzpicture}
\begin{groupplot}[group style={group size= 1 by 1,ylabels at=edge left},width=0.50\textwidth,height=0.25\textwidth]
\nextgroupplot[label style={font=\Large},
tick label style={font=\Large},
legend style={at={($(0,0)+(1cm,1cm)$)},legend columns=3,fill=none,draw=black,anchor=center,align=center,font=\small},
ylabel = {MAE},
legend to name=fred,
mark size=2pt]
\addplot[red!70!black,mark=*] coordinates {(10, 0.316) (40, 0.229) (70, 0.218) (100, 0.233) };
\addplot[brown,mark=star] coordinates {(10, 0.303) (40, 0.220) (70, 0.211) (100, 0.222) };
\addplot[cyan,mark=square*]  coordinates {(10, 0.303) (40, 0.220) (70, 0.211) (100, 0.222) };
\addplot[gray,mark=triangle*] coordinates {(10, 0.303) (40, 0.220) (70, 0.211) (100, 0.222) };
\addplot [blue,mark=diamond*,dashed] coordinates{(10, 0.341) (40, 0.247) (70, 0.240) (100, 0.257) };
\addplot [blue,mark=diamond*] coordinates {(10, 0.204) (40, 0.159) (70, 0.155) (100, 0.164) };

\addlegendentry{CDRec};    
\addlegendentry{DynaMMO};    
\addlegendentry{TRMF};    
\addlegendentry{SVDImp};    
\addlegendentry{\sysname 1D};    
\addlegendentry{\sysname};          
\coordinate (c1) at (rel axis cs:0,1);

\coordinate (c2) at (rel axis cs:1,1);

\end{groupplot}
\coordinate (c3) at ($(c1)!.5!(c2)$);
\node[below] at (c3 |- current bounding box.south)
{\pgfplotslegendfromname{fred}};

\end{tikzpicture}
\caption{MAE (y-axis) for MCAR scenario on JanataHack. X-axis is  percent of time series with missing blocks.}
\label{fig:janta2d}
\end{figure}

\subsubsection{Effect of multidimensional kernel regression}

For this task, we run two variants of our model. The first model dubbed as \sysname1D flattens the multidimensional index of time series by getting rid of the store and product information. The second variant is the proposed model itself which retains the multi-dimensional structure and applies kernel embeddings in two separate spaces. In \sysname\ each time series is associated with two embeddings of size $k$ each. To keep the comparison fair, \sysname1D uses embedding of size $2k$.  Since other methods have no explicit model for multi-dimensional indices, the input is a flattened matrix, similar to \sysname1D.

Figure~\ref{fig:janta2d} shows the performance of the variants compared to the baselines on MCAR for increasing percentage $x$ of number of series with a missing block.  
Observe how in this case too \sysname\ is significantly more accurate than other methods including \sysname1D. 
If each series is small and the number of series is large, there is a greater chance of capturing spurious correlation across series.  In such cases, the external multidimensional structure that \sysname exploits helps to restrict relatedness only among siblings of each dimension.  We expect this difference to get magnified as the number of dimensions increase.

\subsection{Running Time}
\label{subsec:runtime}

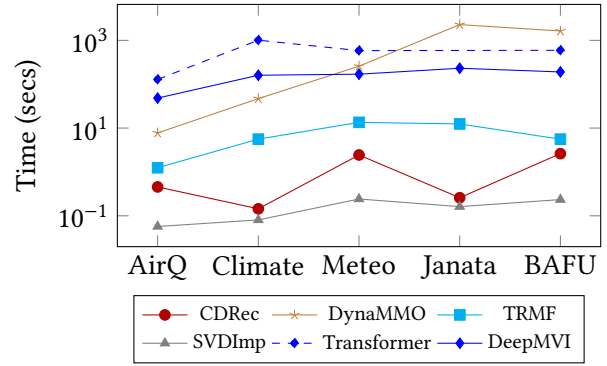
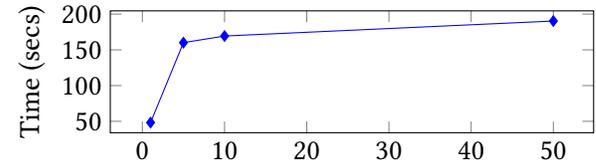
\begin{figure}
\centering
\begin{subfigure}{0.45\textwidth}
\centering
\begin{tikzpicture}[]
\begin{axis}[width=\textwidth,height=0.6\textwidth,
label style={font=\Large},
tick label style={font=\Large},
legend style={at={(0.5,-0.2)},anchor=north,legend columns=3,font=\small},
symbolic x coords={AirQ,Climate,Meteo,Janata,BAFU},
xtick = {AirQ,Climate,Meteo,Janata,BAFU},
ylabel = {Time (secs)},
ymode=log,
mark size=2pt]
\addplot [red!70!black,mark=*] coordinates {(AirQ, 0.455183) (Climate, 0.144344) (Meteo, 2.434875) (Janata,0.258158) (BAFU, 2.612798)};
\addplot [brown,mark=star] coordinates {(AirQ, 7.744843) (Climate, 46.905600) (Meteo, 257.327516) (Janata,2284.452476) (BAFU, 1632.512823)};
\addplot [cyan,mark=square*] coordinates {(AirQ,1.24624490737915) (Climate, 5.59101390838623) (Meteo, 13.5140368938446) (Janata,12.4184200764) (BAFU, 5.59101390838623)};
\addplot [gray,mark=triangle*] coordinates {(AirQ, 0.057099) (Climate, 0.080513) (Meteo, 0.241421) (Janata,0.161895) (BAFU, 0.234143)};
\addplot [dashed,blue,mark=diamond*] coordinates {(AirQ, 128.336043119431) (Climate,1007.940325737) (Meteo, 583.603462696075) (BAFU,592.7444257736206)};
\addplot [blue,mark=diamond*] coordinates {(AirQ, 48.2535245418549) (Climate, 160.085866689682) (Meteo, 169.338786363602) (Janata,231.252223253) (BAFU, 190.30575942993164)};
\addlegendentry{CDRec};    
\addlegendentry{DynaMMO};    
\addlegendentry{TRMF};          
\addlegendentry{SVDImp};          
\addlegendentry{Transformer}; 
\addlegendentry{\sysname}; 
\end{axis}

\end{tikzpicture}
\caption{Absolute Runtime ($s$) on y-axis with dataset arranged in increasing data set size. 
MCAR scenario with all time series having missing blocks (x=100\%).}
\label{fig:runtime}
\end{subfigure}
\begin{subfigure}{0.45\textwidth}
\centering
\begin{tikzpicture}[]
\begin{axis}[width=\textwidth,height=0.4\textwidth,
label style={font=\Large},
tick label style={font=\Large},
ylabel = {Time (secs)},
mark size=2pt]
\addplot [blue,mark=diamond*] coordinates {(1, 48.2535245418549) (5, 160.085866689682) (10, 169.338786363602) (50, 190.30575942993164)};
\end{axis}
\end{tikzpicture}
\caption{Absolute Runtime ($s$) for \sysname\  on y-axis with X-axis showing series length in factor of 1K. The numbers are on real datasets (AirQ,Climate,Meteo,BAFU). Number of time series is 10. More details in Section \ref{subsec:runtime}}
\label{fig:scalability}
\end{subfigure}
\hfill
\caption{Runtime Analysis of \sysname}
\end{figure}

The above experiments have shown that \sysname\ is far superior to existing methods on accuracy of imputation.  One concern with any deep learning based solution is its runtime overheads.  We show in this section that while \sysname\ is slower than existing matrix factorization methods, it is more scalable with TS length and much faster than off-the-shelf deep learning methods.

We present running times on AirQ, Climate, Meteo, BAFU, and JanataHack datasets in Figure~\ref{fig:runtime}. The x-axis shows the datasets ordered by increasing total size
and y-axis is running time in log-scale.  In addition to methods above we also show running time with an off the shelf transformer method.
Matrix factorisation based method like CDRec and SVDImp are much faster than DynaMMO and \sysname.  But compared to the vanilla Transformer our running time is a factor of 2.5 to 7 smaller.
The running time of DynaMMO exceeds the running time of other algorithms by a factor of 1000 and increases substantially with increasing series length, which undermines the accuracy gains it achieves. On the JanataHack dataset, DynaMMO  took 25 mins (1.5e9 $\mu s$) compared to \sysname\ which took just 2.5 mins. 

%
{\color{black}
We next present our numbers on scalability of \sysname\ in Fig \ref{fig:scalability}. The x axis denotes the length of times series in factors of 1K. The points correspond to datasets AirQ, Climate, Meteo and BAFU for 1K, 5K, 10K and 50K respectively. All these datasets have 10 time series. We can see a sub-linear growth of running time with the series length. An intuition behind the same is that our training algorithm learns patterns, which in case of seasonal time series can be learnt by seeing a small number fraction of the series, abstractly one season worth of data.  
}





\subsection{Impact on downstream analytics}
\label{subsec:analysis}

A major motivation for missing value imputation is more accurate data analytics on time series datasets~\cite{Cambronero2017,milo2020automating,kandel2012profiler,Mayfield2010}.  Analytical processing typically involves studying trends of aggregate quantities. 
When some detailed data is missing, a default option is to just ignore the missing data from the aggregate statistic. 
Any MVI method to be of practical significance in data analysis should result in more accurate top-level aggregates than just ignoring missing values.
%
We present a comparison of the different MVI methods on the average over the first dimension so the result is a $n-1$ dimensional aggregated time series.  Except in JanataHack and M5, this results in a single averaged time series.

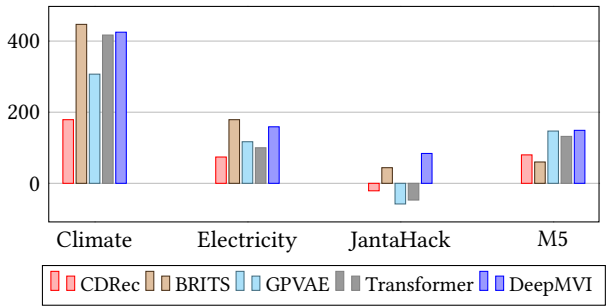
\begin{figure}
\centering
\begin{tikzpicture}
\begin{groupplot}[group style={group size= 1 by 1},width=0.5\textwidth,height=0.25\textwidth]
\nextgroupplot[ybar=1pt, 
symbolic x coords={Climate,Electricity,JantaHack,M5},
xtick={Climate,Electricity,JantaHack,M5},
scaled y ticks = false,
legend pos = north west, 
tickwidth=0pt,
ymajorgrids=true,
legend style={at={(0.5,-0.2)},anchor=north,legend columns=5,font=\small},
bar width=4pt]
\addplot+[red, fill=red!30!white] coordinates {(Climate, 179) (Electricity,74) (JantaHack, -21) (M5, 80) };
\addplot+[brown!40!black,fill=brown!50!white] coordinates {(Climate, 447) (Electricity, 179) (JantaHack, 44) (M5, 60) };
\addplot+[cyan!40!black, fill=cyan!30!white] coordinates {(Climate, 307) (Electricity, 117) (JantaHack, -58) (M5, 147) };
\addplot+[gray, fill=black!40!white] coordinates {(Climate, 417) (Electricity, 100) (JantaHack, -47) (M5, 132) };
\addplot+[blue, fill=blue!40!white] coordinates {(Climate, 425) (Electricity, 159) (JantaHack, 84) (M5, 149) };

\legend{CDRec,BRITS,GPVAE,Transformer,\sysname};
\coordinate (c1) at (rel axis cs:0,1);

\end{groupplot}

\end{tikzpicture}
\caption{Difference between Mean Absolute Error of \DiscardT\ and each algorithm is on y-axis, and four other datasets on x-axis.}
\label{fig:barplots}
\end{figure}

{\color{black}
Apart from computing the above statistic on the imputation output by various algorithms, we also compute this statistic with just the missing values in the set dropped from the average. We call this the \DiscardT\ method.
%
We consider four datasets: Climate, Electricity, JanntaHack, and M5, each in MCAR with $100\%$ of the time series containing missing values. On each of these datasets, we first compute the aggregate statistic using true values. For Climate and Electricity, this returns a single series with value at time $t$ as the average of values at all series at time $t$. For JantaHack, we average over 76 stores resulting in average sales of 28 products. Similarly on M5, we average over 10 stores giving us sales of 106 items. Next we compute the aggregate statistic with missing values imputed by five algorithms: CDRec, BRITS, GPVAE, Transformer, and DeepMVI. We compute MAE between aggregate with imputed values and aggregate over true values.

In Figure~\ref{fig:barplots}, we report the difference between MAE of \DiscardT\ and MAE of the algorithm.  We see that on the JanataHack dataset, three existing imputation method CDRec, GPVAE, and Transformer provide worse results than just dropping the missing value in computing the aggregate statistic.  In contrast \sysname\ provides gains over this default in all cases. Also, it is overall better than existing methods, particularly on the multidimensional datasets.  This illustrates the impact of \sysname\ on downstream data analytics.
}



\section{Conclusion and Future Work}
In this paper, we propose \sysname, a deep learning method for missing value imputation in multi-dimensional time-series data. \sysname\ combines within-series signals using a novel temporal transformer, across-series signals using a multidimensional kernel regression, and local fine-grained signals.  The network parameters are carefully selected to be trainable across wide ranges of data sizes, data characteristics, and missing block pattern in the data.   

We extensively evaluate \sysname\ on ten datasets, against comparing seven conventional and three deep learning methods, and with five missing-value scenarios. \sysname\ achieves up to 70\% error reduction compared to state of the art methods. Our method is up to 50\% more accurate and six times faster than using off-the-shelf neural sequence models.
%
We also justify our module choices by comparing \sysname\ with its variants. We show that \sysname's performance on downstream analytics tasks is better than dropping the cells with missing values as well as existing methods.

Future work in this area includes applying our neural architecture to other time-series tasks including forecasting.



\bibliographystyle{ACM-Reference-Format}
\bibliography{sample}

\end{document}